\documentclass[runningheads]{llncs}

\usepackage{bibunits}

\defaultbibliographystyle{splncs04}
\usepackage{eccvabbrv}

\usepackage{comment}
\usepackage{xspace}
\newcommand{\GFM}{GFM\xspace}
\newcommand{\GFMs}{GFMs\xspace}
\usepackage[table]{xcolor}
\definecolor{best}{RGB}{198,239,206}
\definecolor{second}{RGB}{255,242,204}
\usepackage{xcolor}
\usepackage{tcolorbox}
\usepackage[percent]{overpic}
\usepackage{multirow}
\usepackage{tikz}
\usetikzlibrary{arrows.meta, positioning, calc}

\newcommand{\removeButImportant}[1]{}

\definecolor{takeawaybg}{RGB}{245,247,250}
\definecolor{takeawayborder}{RGB}{180,190,200}

\newtcolorbox{takeawaybox}{
    colback=takeawaybg,
    colframe=takeawayborder,
    boxrule=0.5pt,
    arc=2pt,
    left=6pt,
    right=6pt,
    top=6pt,
    bottom=6pt
}

\usepackage{graphicx}

\usepackage{caption}
\usepackage{subcaption}

\usepackage{booktabs}
\graphicspath{./figures}
\renewcommand{\eg}{{e.g.,}\xspace}
\renewcommand{\ie}{{i.e.,}\xspace}
\usepackage[accsupp]{axessibility}
\newcommand{\groot}{{GR00T}\xspace}

\usepackage{hyperref}
\usepackage{bm}
\usepackage{orcidlink}

\newcommand{\calI}{{\cal I}}

\newcommand{\calL}{{\cal L}}

\newcommand{\calR}{{\cal R}}

\newcommand{\myParagraph}[1]{{\bf #1.}\xspace}

\newcommand{\M}[1]{{\bm #1}} 
\renewcommand{\boldsymbol}[1]{{\bm{#1}}}

\newcommand{\tran}{^{\mathsf{T}}}

\newcommand{\e}{{\mathrm e}}

\newcommand{\Real}[1]{ { {\mathbb R}^{#1} } }

\newcommand{\MA}{\M{A}}

\newcommand{\MG}{\M{G}}

\newcommand{\MR}{\M{R}}

\newcommand{\MV}{\M{V}}

\newcommand{\ML}{\M{L}}

\newcommand{\MT}{\M{T}}
\newcommand{\MX}{\M{X}}
\newcommand{\MY}{\M{Y}}
\newcommand{\MW}{\M{W}}
\newcommand{\MZ}{\M{Z}}

\newcommand{\va}{\boldsymbol{a}} 
 
\newcommand{\vb}{\boldsymbol{b}}

\newcommand{\vn}{\boldsymbol{n}}

\newcommand{\vxx}{\boldsymbol{x}}

\begin{document}

\title{Understanding the Impact of Geometric Foundation Models
on Vision-Language-Action Models}

\titlerunning{Understanding the Impact of Geometric Foundation Models
on VLAs}

\author{Yurou Yang\inst{1} \and
Muyuan Lin\inst{1} \and
Roberto Martin-Martin\inst{1,2} \and
Martin Labrie\inst{1} \and
Shreekant Gayaka\inst{1} \and
Cheng-Hao Kuo\inst{1} \and
Luca Carlone\inst{1,3}
}

\authorrunning{Yang et al.}

\institute{Amazon Personal Robotics Group\inst{1}
\qquad University of Texas at Austin\inst{2}
\qquad Massachusetts Institute of Technology\inst{3}}

\maketitle


\vspace*{-1em}
\begin{abstract}
Recent work explores new opportunities at the intersection of vision-language-action models (VLAs) and geometric foundation models (\GFMs) for 3D reconstruction, such as VGGT. While the resulting \textit{geometric} VLAs often show improved performance, it remains unclear (i) if modern VLAs already have sufficient geometric understanding to start with,
(ii)~what is the best architecture to inject geometric understanding into a VLA, and (iii) 
what is the effect of other design choices that affect {geometric VLAs}.
In this paper we provide a rigorous experimental analysis to shed light on these questions, for a specific choice of VLA (\groot-N1.5) and \GFM (VGGT). 
Our first contribution is to formalize prior work's intuition that current VLAs lack geometric understanding, by providing a rigorous analysis based on linear probing. The analysis quantifies, for the first time, the ``geometric gap'' between VLAs and \GFMs.
Our second contribution is to identify and compare different strategies to bridge \GFMs with VLAs.
We implement three different architectures, which differ in the way they inject geometry in the VLA, while keeping low-level implementation details as similar as possible, to ensure a fair comparison.
Finally, we analyze the impact of non-architectural choices (\eg training data, number of cameras, reconstruction quality) on the performance of the geometric VLAs. 
\end{abstract}


\section{Introduction}
Vision-Language-Action models (VLAs) have recently been at the center stage of robotics research for their ability to enable generalist manipulation behaviors from a modest number of expert demonstrations. 
These models have enabled successful manipulation skills from text instructions (\eg ``pick up the red mug and put it in the sink'') and are steadily 
permeating from academic labs to industrial applications (\eg~\cite{figureai_helix_2025,tesla_ai_2026,pi_website_2026}).
Typical VLAs are composed of two main building blocks: a \emph{vision-language model} (VLM), which parses input images collected by the robot and language instructions from the user, and an \emph{action expert}, which takes the output of the VLM and computes a suitable action (or chunk of actions) for the robot to execute. Intuitively, the VLM helps the robot correlate the user's instructions (``pick up the red mug'') with relevant pixels in the image, while the action expert is in charge of establishing the best course of action depending on the robot embodiment and the VLM understanding.
%
%
While the combination of VLMs and action expert has demonstrated to be a strategic and powerful architectural choice, VLMs have been observed to be poor at spatial understanding~\cite{Chen24cvpr-spatialVLM,cai2025spatialbot}. 
At the same time, spatial understanding is intrinsically related to manipulation, where distances and geometry make the difference between a successful physical interaction and a failed one. 
This insight triggered work on \emph{3D} VLAs, where the VLA policy is augmented with depth information from an RGB-D camera~\cite{Li25arxiv-pointVLA,Qu25arxiv-spatialVLA,Zhen24arxiv-3DVLA,Chen24cvpr-SUGAR,Goyal24arxiv-RVT2,Jia24arxiv-Lift3D}. 

Parallel work in computer vision has pioneered the use of feed-forward transformer architectures for 3D reconstruction. Contrary to traditional SLAM and Structure from Motion approaches, which decouple the problem into multiple stages (\eg feature extraction, matching, bundle adjustment, dense 3D reconstruction), these new \emph{geometric foundation models} (\GFM) reconstruct camera poses, depth maps, and point maps of the observed scene from a collection of images (\eg~\cite{Wang25arxiv-vggt,Fang25arxiv-dens3r,Peng25arxiv-omnivggt,Wang25arxiv-amb3r,Rojas25arxiv-hamst3r}). The advantage of these approaches lies in their simplicity, their zero-shot generalization, and their ability to work with uncalibrated cameras. 
These new methods have created new opportunities to enhance VLAs with geometric understanding without altering their input data. 
This is particularly important since VLAs are typically trained on RGB inputs 
and \GFMs open the door to adding 3D information without requiring additional data collection. 
These opportunities have been explored in a number of very recent works~\cite{Lin25arxiv-Evo0VLA,Abouzeid25arxiv-GeoAwareVLA,Ni25arxiv-VO-DP,Ge25arxiv-VGGTDP,Zhang25arxiv-FALCONSpatialToActions,Li25arxiv-SpatialForcing}, which report promising results.

Despite the quick progress in combining VLAs and \GFMs in a so-called \textit{geometric VLA}, related work leaves several questions unanswered. First of all, it remains unclear if VLAs already have sufficient geometric understanding and how to quantify a potential gap in 3D understanding.
Second, related works propose disparate architectures to inject 3D information from \GFMs into VLAs and it is unclear how these architectures compare or what is the insight behind one architecture being better than another. Finally, it is unclear how external factors related to geometry, including number of cameras, size of the training set, and reconstruction quality, impact performance of geometric VLAs.

\begin{figure*}[t]
\centering
\setlength{\tabcolsep}{2pt} 
\renewcommand{\arraystretch}{1.0}

\begin{tabular}{c|c}
    \begin{minipage}{0.47\textwidth}
        \centering
        \begin{overpic}[width=\linewidth, trim=80mm 75mm 125mm 70mm, clip]{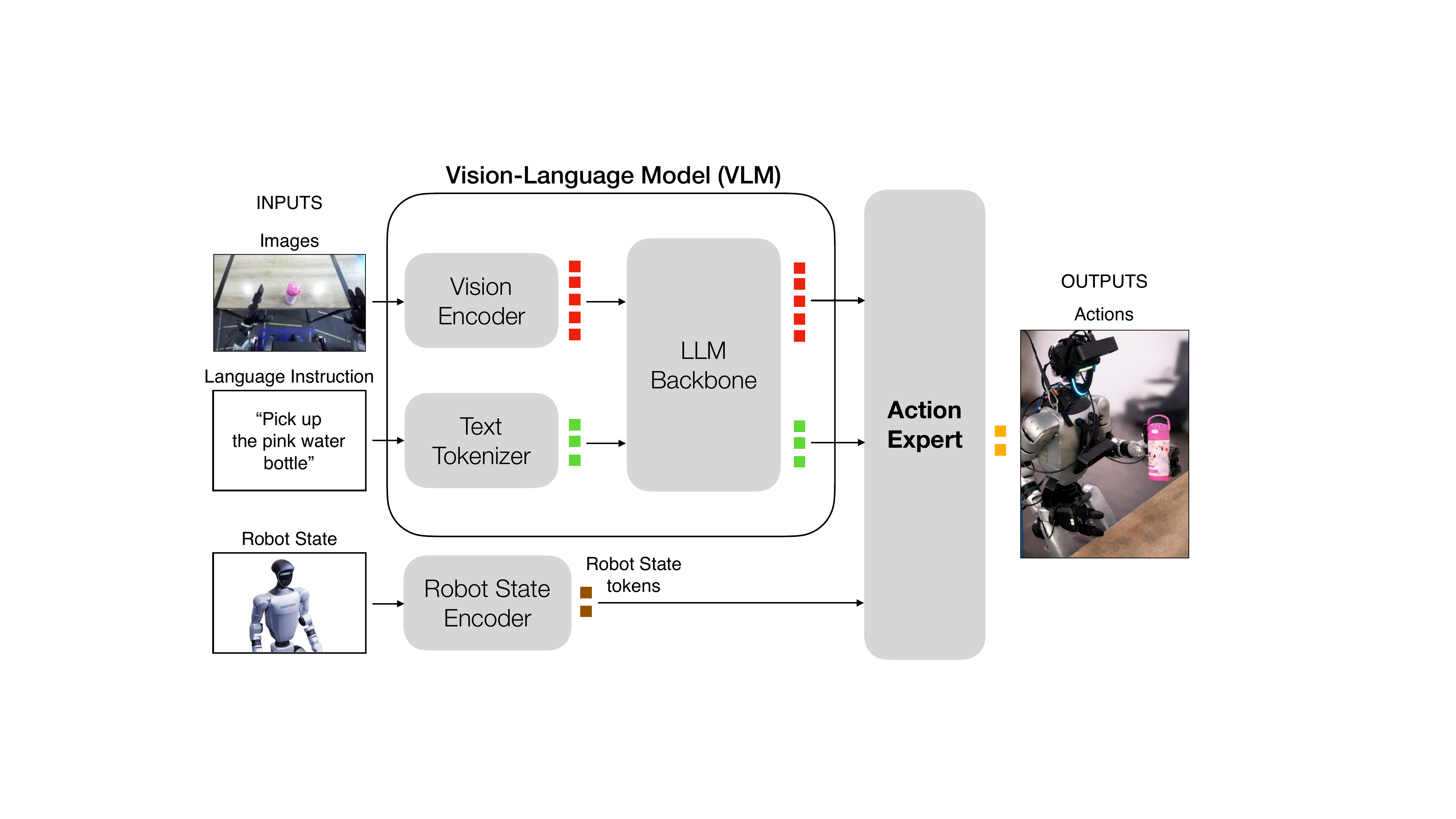}
        \put(38,40){{\tiny $\MV_e$}}
        \put(38,25){{\tiny $\ML_e$}}
        \put(60,40){{\tiny $\MV_l$}}
        \put(60,25){{\tiny $\ML_l$}}
        \put(42,2){{\tiny $\MR$}}
    \end{overpic}
        \vspace{0mm}
        
        {(a)} Standard VLA
        \vspace{2mm}
    \end{minipage}
    &
    \begin{minipage}{0.47\textwidth}
        \centering
        \includegraphics[width=\linewidth, trim=80mm 75mm 125mm 40mm, clip]{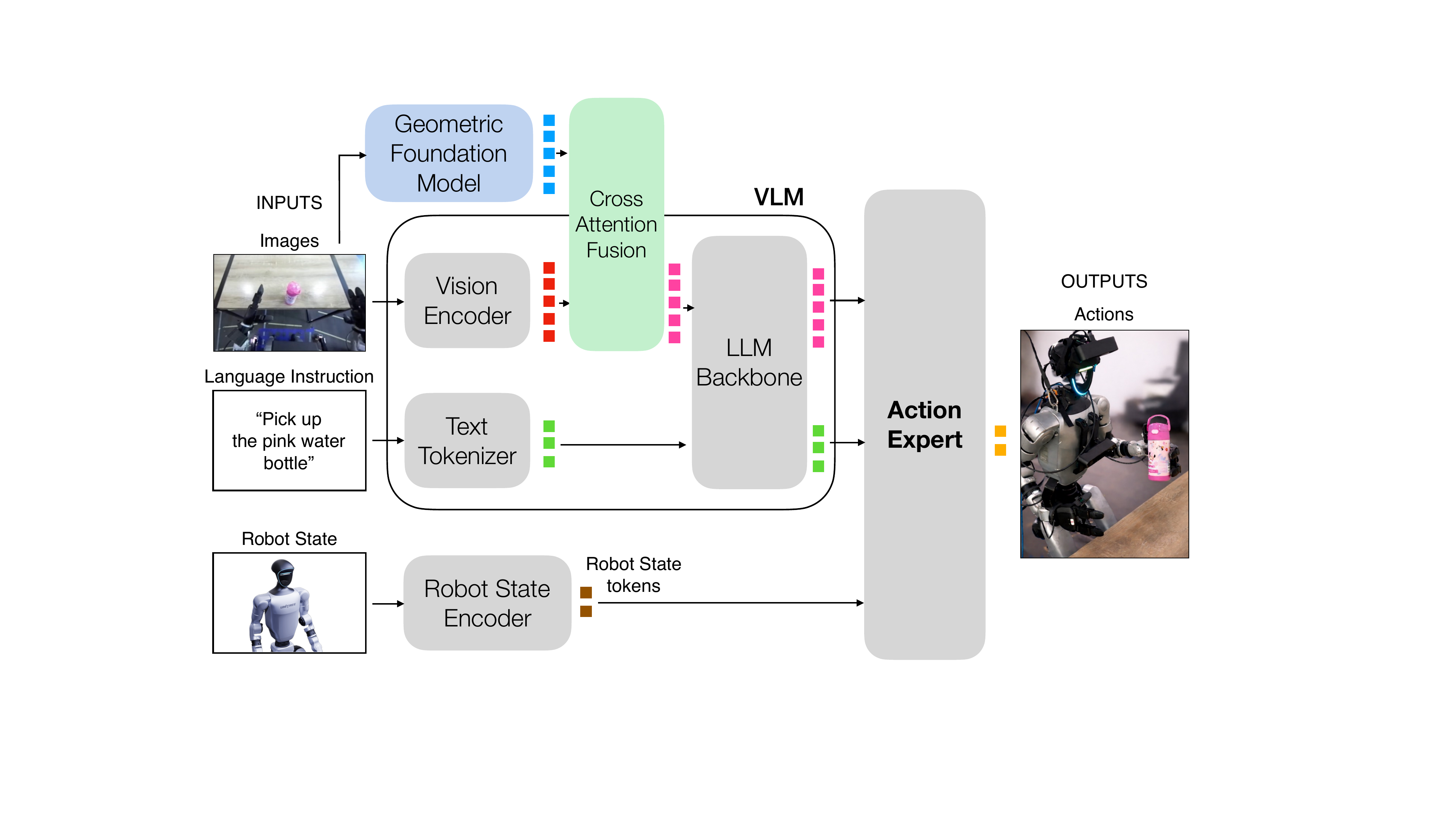}
        
        {(b)} Early Fusion
        \vspace{2mm}
    \end{minipage}
    \\

    \hline

    \begin{minipage}{0.47\textwidth}
        \centering
        \includegraphics[width=\linewidth, trim=80mm 75mm 125mm 40mm, clip]{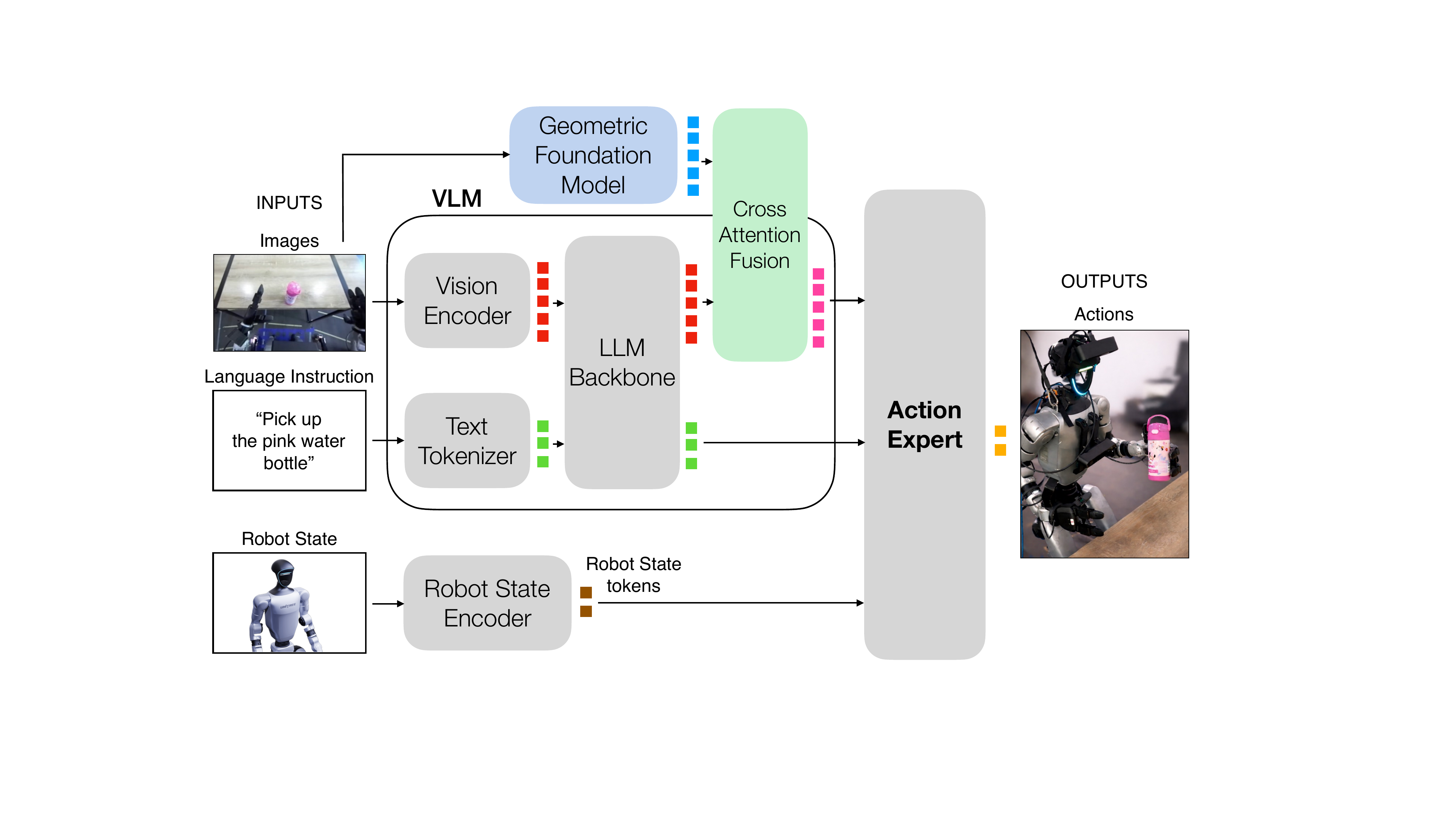}
        
        {(c)} Late Fusion
    \end{minipage}
    &
    \begin{minipage}{0.47\textwidth}
        \centering
        \includegraphics[width=\linewidth, trim=80mm 75mm 125mm 40mm, clip]{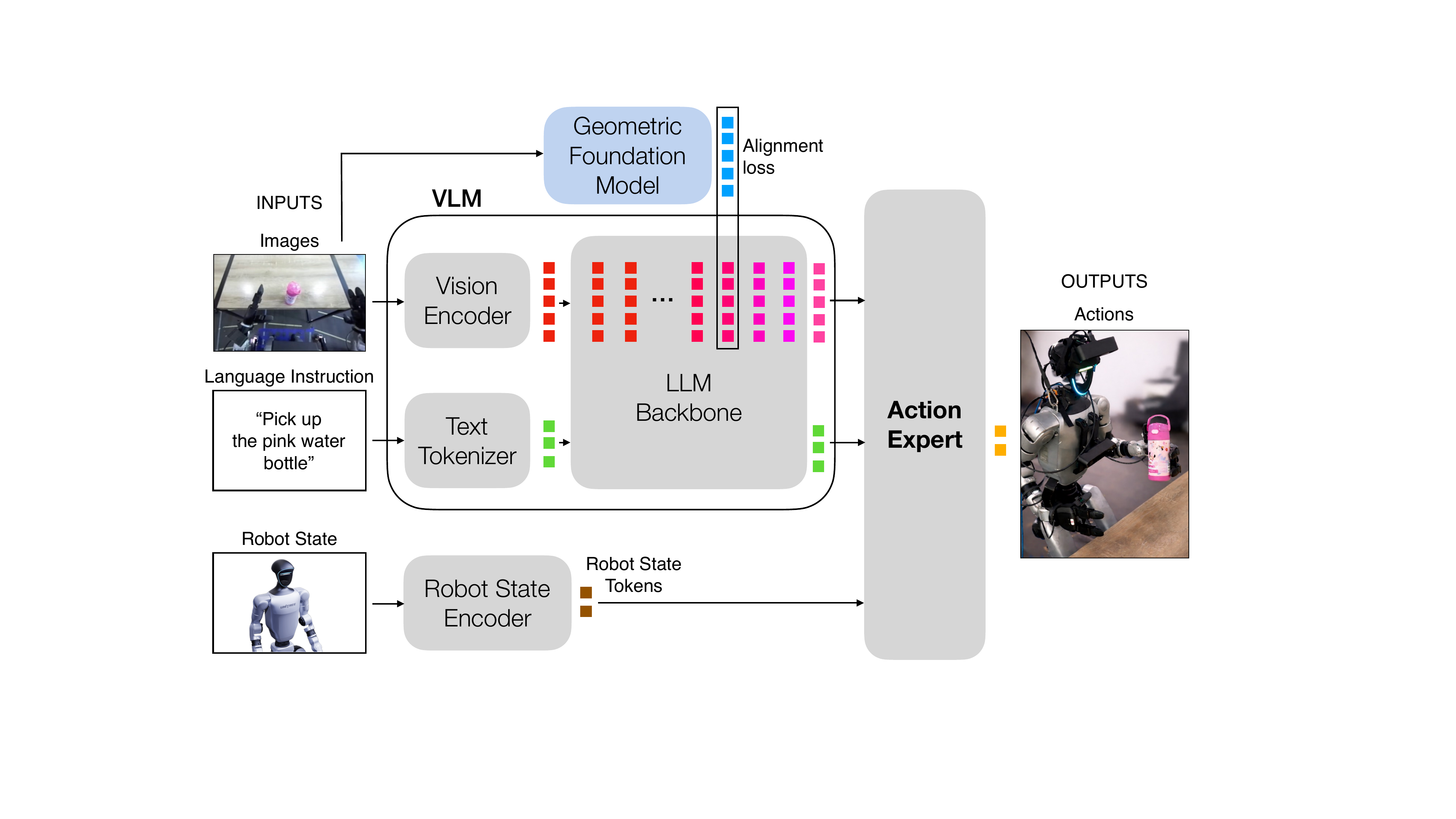}
        
        {(d)} Spatial Forcing~\cite{Li25arxiv-SpatialForcing}
    \end{minipage}
\end{tabular}

\caption{(a) Standard VLA architecture. (b)-(d) Key strategies to inject tokens from a geometric foundation model (\eg VGGT) into the VLA.}
\label{fig:architectures}

\end{figure*}

\myParagraph{Contributions} In this paper we provide a rigorous experimental analysis to shed light on the open questions above, for a specific choice of VLA (\groot-N1.5) and \GFM (VGGT). 
Our first contribution (Section~\ref{sec:architectures}) is to {\bf identify and compare different strategies to bridge \GFMs and VLAs}. 
Towards this goal, we analyze related work and identify three different strategies, summarized in Fig.~\ref{fig:architectures}.
We provide prototype implementations of models using the three different strategies, while keeping low-level implementation details as similar as possible (to ensure an apple-to-apple comparison). 
%
Our second contribution (Section~\ref{sec:probing}) is to formalize related work's intuition that current VLAs lack geometric understanding, by providing a {\bf rigorous analysis based on linear probing}. The probing assesses the VLA ability (more specifically, the ability of the VLM within the VLA) to predict dense depth information from images.
The analysis quantifies, for the first time, the ``geometric gap'' between VLAs and \GFMs, rigorously grounding observations in related work~\cite{Li25arxiv-SpatialForcing}. Our analysis also provides the surprising finding that a pretrained LLM can predict dense depth information if it is fed with tokens from a \GFM.
Our final contribution (Section~\ref{sec:factors}) is to {\bf analyze the effect of key design choices that affect geometric VLAs}. 
Relevant factors include both architectural aspects (such as the choice of fusion strategy discussed above), as well as non-architectural aspects (such as the number of cameras used by the VLA, the size of the training data, and the quality of the reconstruction from the \GFM).
We perform our analysis on popular simulation benchmarks, including RoboCasa~\cite{Nasiriany24rss-robocasa} and LIBERO~\cite{Liu23arxiv-libero}, and we 
also provide an evaluation on real data, using the Unitree G1 humanoid.






\section{Related Work}\label{sec:related_work}

\myParagraph{Vision-Language-Action Models (VLAs) for Manipulation} 
VLAs map visual observations and natural language instructions to robot actions by extending VLMs with a policy or action head~\cite{black2024pi0,intelligence2025pi05,bjorck2025gr00t,brohan2022rt,driess2023palm}. 
These models inherit the semantic grounding and generalization capabilities of large VLMs, enabling robots to perform diverse manipulation tasks from relatively limited demonstration data. However, despite their impressive generalization capabilities, VLAs inherit a key limitation from VLMs: they primarily operate on RGB inputs and focus on semantic understanding~\cite{ahn2022can} rather than explicit spatial reasoning.
Prior work has shown that vision-language models often struggle with geometric reasoning tasks such as estimating distances, relative pose, or object placement constraints~\cite{Qu25arxiv-spatialVLA,cai2025spatialbot}.
While recent work attempts to improve spatial reasoning through architectural modifications or large-scale multimodal pretraining~\cite{Qu25arxiv-spatialVLA,Zhen24arxiv-3DVLA,Zhu24arxiv-LLaVA3D,comanici2025gemini}, it remains unclear whether these approaches provide the level of geometric understanding required for efficient robotic manipulation.


\myParagraph{Geometric Foundation Models} 
Recent work in computer vision has introduced \emph{geometric foundation models} (GFMs), which learn to infer 3D scene structure directly from images using a single feed-forward transformer-based architecture~\cite{Wang24cvpr-dust3r,Leroy24eccv-mast3r}.
Models such as the Visual Geometry Grounded Transformer (VGGT)~\cite{Wang25arxiv-vggt} predict camera poses, depth maps, and dense point maps from one or multiple views without the multi-stage pipelines used in traditional SLAM or structure-from-motion systems.
Subsequent work has further developed these models with improved geometric consistency by jointly predicting correlated geometric quantities such as depth, normals, and point maps~\cite{Wang24cvpr-dust3r}, incorporating additional modalities or supervision to enable metric reconstruction and multi-view consistency~\cite{Keetha25arxiv-mapanything,Wang25arxiv-amb3r}, or extending these architectures to dynamic scenes and semantic understanding, enabling unified feed-forward modeling of geometry, semantics, and motion~\cite{Zust25arxiv-panst3r,Rojas25arxiv-hamst3r,Hu25arxiv-vggt4d}.
Together, these models provide powerful geometric representations that generalize across environments and can be queried without scene-specific training. These techniques have recently been shown to improve spatial understanding in VLMs~\cite{Zheng25neurips-3DMLLM,Wu25neurips-spatialMLLM,Hu25cvpr-G2VLM}.

\myParagraph{Injecting Geometry into VLAs} 
Several works augment VLAs with geometric information to address the limited spatial reasoning capabilities of traditional VLM-based policies.
Some approaches incorporate explicit geometric inputs such as depth maps or point clouds, for example PointVLA~\cite{Li25arxiv-pointVLA} and RVT-2~\cite{Goyal24arxiv-RVT2}.
Other works attempt to learn spatial structure within the policy itself, through 3D feature learning, affordance reconstruction, or spatial memory mechanisms~\cite{Chen24cvpr-SUGAR,Jia24arxiv-Lift3D,Steiner25arxiv-mindmapSpatialMemory}.
More recently, geometric foundation models have been used to inject geometric context into VLAs, either by fusing geometric tokens with visual representations~\cite{Lin25arxiv-Evo0VLA,Abouzeid25arxiv-GeoAwareVLA,Ni25arxiv-VO-DP,Ge25arxiv-VGGTDP} or by aligning vision-language representations with geometric features during training~\cite{Li25arxiv-SpatialForcing}.
Despite promising results, existing approaches differ substantially in how geometry is integrated, making it difficult to understand which architectural choices are most effective.
In this work, we provide a systematic analysis of several strategies for integrating geometric foundation models into modern VLAs.

\section{Bridging VLAs and Geometric Foundation Models} 
\label{sec:architectures}

This section reviews the typical VLA architecture and discusses ways to integrate \GFMs in 
a VLA (Section~\ref{sec:architectures-overview}). Then, we describe our implementation of the ``fusion module'' that fuses VLA tokens with tokens from the \GFM (Section~\ref{sec:crossattn}).

\subsection{VLA Architecture and Key Strategies to Inject Geometry} 
\label{sec:architectures-overview}

\myParagraph{VLA Architecture} A typical VLA is shown in Fig.~\ref{fig:architectures}(a). 
The VLA takes three inputs: a set of images $\calI$ (\eg captured by the robot cameras at the current time $t$), a language instruction $\calL$ (\eg ``pick up the red mug''), and robot state information $\calR$ (\eg  joint angles).
The inputs are first processed by specific encoders 
that produce visual ($\MV_e$), language ($\ML_e$), and robot state tokens ($\MR$), respectively.
Visual and language tokens are then passed to the LLM backbone inside the VLM
that
produces new visual tokens ($\MV_l$) and language tokens ($\ML_l$). These tokens, together with the robot state tokens $\MR$, are passed to the action expert, which computes the set of actions for the next $T$ time steps, $\va_{0:T}$. 

The VLA is trained from demonstrations, given tuples $(\calI,\calL, \calR, \va_{0:T})$, 
including both input data $(\calI,\calL, \calR)$ and the desired action sequence, $\va_{0:T}$, given in the demonstration. 
While different architectures differ in the choice of the VLM, the action expert, and 
training objective, a popular choice is to use a diffusion policy formulation~\cite{Chi23rss-diffusionpolicy}, 
where the action expert integrates a diffusion transformer.
In a diffusion policy formulation, one iteratively adds Gaussian noise to the clean action sequence, $\va_{0:T}$, to obtain a noisy sequence $\mathbf{a}^k$, $k=1,\ldots, N$, such that:
\begin{equation}
\mathbf{a}^k = {\alpha_k}\va_{0:T}
+ {(1-\alpha_k)}\boldsymbol{\epsilon}, \quad
\boldsymbol{\epsilon} \sim \mathcal{N}(0, \mathbf{I}).
\end{equation}

The diffusion policy, parametrized by weights $\theta$, is then trained with a flow-matching loss, which 
teaches the policy how to map back random noise samples to plausible actions.
Concretely, the policy predicts the denoising field $v_\theta(\mathbf{a}^k, \MT, k)$, 
conditioned on the visual, text, and robot state tokens, $\MT = (\MV_l,\ML_l,\MR)$, and is trained using the following flow-matching loss:
\begin{equation}
\mathcal{L}_{\text{diff}} =
\mathbb{E}_{k} 
\left[
\left\|
v_\theta(\mathbf{a}^k, \MT, k) - (\boldsymbol{\epsilon}
- \va_{0:T})
\right\|_2^2
\right].
\end{equation}
The predicted robot actions are then obtained via (Euler) integration of the field $v_\theta(\mathbf{a}^k,\MT, k)$; see~\cite{black2024pi0,bjorck2025gr00t} for details and popular instantiations of these ideas.

\myParagraph{Early Fusion} Now let us consider a first strategy to inject \GFM tokens into the baseline VLA described above. This first strategy, that we call ``Early Fusion'', is illustrated in Fig.~\ref{fig:architectures}(b).
More specifically, the \GFM takes the same images $\calI$ as the VLA and produces tokens $\MG$; in our experiments, we use the VGGT backbone to produce these tokens.
Then, visual tokens $\MV_e$ from the VLA and geometric tokens $\MG$ from the \GFM are fused together to produce new tokens ${\tt fuse}(\MV_e, \MG)$, which are passed to the LLM in lieu of the tokens from the vision encoder. 
Intuitively, the Early Fusion strategy treats the \GFM as an additional encoder. 
This approach has been used in early papers combining VGGT with VLAs~\cite{Lin25arxiv-Evo0VLA}; moreover, it resembles injection strategies used to enhance spatial understanding in VLMs~\cite{Zheng25neurips-3DMLLM}. We implement the fusion ${\tt fuse}(\MV_e, \MG)$ as a cross-attention layer with attention gating, as discussed in Section~\ref{sec:crossattn} below.
We remark that the use of attention gating was not explicitly mentioned in~\cite{Lin25arxiv-Evo0VLA,Zheng25neurips-3DMLLM}, but we found this aspect to be crucial to the effectiveness of this strategy (Appendix~\ref{sec:ablations}). 

\myParagraph{Late Fusion} The second strategy, that we call ``Late Fusion'', is illustrated in Fig.~\ref{fig:architectures}(c).
Again, the \GFM takes the same images $\calI$ as the VLA and produces geometric tokens, $\MG$. 
However, in this case the geometric tokens are fused with the visual tokens outputted by the LLM, namely $\MV_l$; the fused tokens, ${\tt fuse}(\MV_l, \MG)$, are used as input by the action expert instead of the original LLM tokens, $\MV_l$. 
Intuitively, the Late Fusion strategy attempts at enriching the VLM output with more geometric information before passing it to the action expert. 
This approach is conceptually similar to~\cite{Zhang25arxiv-FALCONSpatialToActions}, but with a simplified architecture. As before, we implement the fusion ${\tt fuse}(\MV_l, \MG)$ as a cross-attention layer with attention gating, to keep the implementation as close as possible across architectures.

\myParagraph{Spatial Forcing} The third strategy is Spatial Forcing~\cite{Li25arxiv-SpatialForcing} (see Fig.~\ref{fig:architectures}(d)).
In this case, the architecture itself is the same as the original VLA 
with the main difference being that, at training time, spatial forcing adds an \emph{alignment loss} to encourage internal tokens of the LLM to align (as measured by cosine similarity) with the \GFM tokens. This facilitates the VLM to retain geometric information that could potentially be used by the action expert.

\begin{remark}[Our evaluation focuses on \groot and VGGT]
At the time of preparation of this manuscript, no open-source code was available for approaches following the Early Fusion and Late Fusion strategy. Therefore, we developed prototype implementations for both strategies, building on top of the \groot-N1.5 VLA~\cite{nvidia_gr00t_n1_5_2025}.
For consistency, we also implemented Spatial Forcing~\cite{Li25arxiv-SpatialForcing} using  \groot: while open-source code is available for Spatial Forcing, it is based on OpenVLA~\cite{Kim24arxiv-openVLA} and $\pi_0$~\cite{black2024pi0}; we adapted to \groot for a fair comparison. 
In all tests, we used VGGT as the geometric foundation model.
\end{remark}

\subsection{Cross-Attention Fusion}
\label{sec:crossattn}
In this section, we briefly discuss the implementation of the fusion module ${\tt fuse}(\cdot,\cdot)$ mentioned in Section~\ref{sec:architectures-overview}. We adopt a similar implementation for both the Early Fusion and the Late Fusion strategy. Let us denote with $\MX \in \Real{P_x \times D_x}$ the VLA visual tokens, where $P_x$ denotes the number of tokens (this typically depends on the number of patches the encoder subdivides the images in) and $D_x$ is the size of the feature associated to each token;
for the Early Fusion, we set $\MX = \MV_e$, while for the Late Fusion we set $\MX = \MV_l$. Also, denote with $\MG \in \Real{P_g \times D_g}$ the tokens of the geometric foundation model. 
Below we show how to obtain fused tokens $\tilde{\MX} = {\tt fuse}(\MX,\MG)$ using a cross-attention mechanism with attention gating, such that $\tilde{\MX} \in \Real{P_x \times D_x}$ has the same size of the VLA tokens, hence remaining compatible with the rest of the architecture.

We follow a standard cross-attention implementation, where tokens are first projected onto a common space using learned projection matrices ${\MW}_Q, {\MW}_K ,{\MW}_V$:

\vspace{-4mm}
\begin{align}
\label{eq:QKV}
\mathbf{Q} &= \MX \; {\MW}_Q \in \mathbb{R}^{P_x \times d}, \quad
\mathbf{K} &= \MG \; {\MW}_K \in \mathbb{R}^{P_g \times d}, \quad
\mathbf{V} &= \MG \; {\MW}_V \in \mathbb{R}^{P_g \times d}.
\end{align}

Then, cross-attention is computed as follows:
\begin{equation}
\label{eq:crossAtt}
\MY =
\text{softmax}
\left(
\frac{\mathbf{Q}\mathbf{K}^\top}{\sqrt{d}}
\right)\mathbf{V} \in \Real{P_x \times d}
\end{equation}
and the result is projected back to the VLA feature space using a learned projection matrix $\MW_O$:
\begin{equation}
\label{eq:projWO}
\MZ = \MY \; \MW_O
\in \Real{P_x \times D_x}.
\end{equation}
Finally, we apply a learned attention gate $\MA$ to the cross-attention results and 
use the gated results as a residual correction term on the VLA tokens:
\begin{equation}
\label{eq:gate}
\tilde{\MX} =
\MX + \MA \odot \MZ
\end{equation}
where $\odot$ denotes the component-wise (Hadamard) product.
The gate is initialized close to zero, such that the geometric tokens are gradually added to the VLA tokens; 
this aspect is important since the rest of the architecture is pretrained without the geometric tokens, and 
 introducing geometric tokens too abruptly prevents the action expert from correctly using them (Appendix~\ref{sec:ablations}).
In our implementation, we follow a standard practice and parameterize the linear projections ${\MW}_Q, {\MW}_K ,{\MW}_V, \MW_O$ as low-ranking (LoRA) layers~\cite{hulora} 
and add positional encodings to the VLA and \GFM tokens to retain the spatial arrangement of the corresponding tokens.
We refer the reader to Appendix~\ref{sec:architecture_details} for more details.


\section{Probing Geometric Understanding in VLAs} 
\label{sec:probing}

\begin{takeawaybox}
\textbf{Key Takeaway.}
GR00T-N1.5 is not able to retain depth information, which is already lost before the output of the VLM.
Late fusion, as expected, can inject depth understanding.
Surprisingly, even Early Fusion can inject depth understanding, without requiring finetuning of GR00T's VLM.
\end{takeawaybox}

Before investigating the impact of injecting geometric tokens into a VLA, we ask a basic question: do VLAs already understand geometry? We rigorously answer this question using a technique known as \emph{linear probing}. 
In linear probing, one attaches a simple MLP (the linear probe) to a layer of a frozen network, and then trains the MLP to solve a task (in our case, monocular depth estimation) to probe the information content of that layer: intuitively if the layer contains enough information (\eg about geometry), then the task can be completed successfully, otherwise it cannot. Qualitative probing results have been reported in related work (\eg~\cite{Li25arxiv-SpatialForcing}), but we provide the first \emph{quantitative} evaluation of the geometric understanding gap between VLAs and geometric foundation models.

\myParagraph{Probing Setup} We train the linear probe for 10 epochs on the NYU Depth V2 dataset (we use the version of the dataset from~\cite{Wofk19icra-fastDepth}). The dataset includes pairs of RGB and depth images, including 24,000 training pairs, and 645 pairs for validation. For probing purposes, we use a Scale-Invariant Logarithmic (SILog) loss following standard practice~\cite{Yang24arxiv-depthAnything2}.  To evaluate probing performance, we compute the RMSE depth error ($\downarrow$),  as well as the $\delta_1$ score ($\uparrow$), which measures the fraction of pixels with predictions within $25\%$ of the ground truth depth. 

\begin{figure*}[t]
\centering
\vspace{0pt}
\captionsetup{width=\linewidth}

\noindent
\begin{minipage}[t]{0.5\textwidth}
\centering
\vspace{0pt}
\small
\captionsetup{width=\linewidth}

\resizebox{\linewidth}{!}{
\begin{tabular}{lcc}
\hline
 & RMSE [m] ($\downarrow$) & $\delta_1$ ($\uparrow$) \\
\hline
GR00T - vision encoder probe & 0.92 & 0.51 \\
GR00T - VLM probe            & 0.73 & 0.63 \\
VGGT probe                   & {\bf 0.41} & {\bf 0.89} \\
\hline
Early Fusion probe           & 0.44 & 0.88 \\
Late Fusion probe            & 0.45 & 0.87 \\
\hline
\end{tabular}
}
\vspace{-2mm}
\captionof{table}{RMSE depth errors and $\delta_1$ scores for the linear probing experiments.}
\label{tab:probe_comparison}
\end{minipage}
\hfill
\begin{minipage}[t]{0.45\textwidth}
\centering
\vspace{0pt}
\captionsetup{width=\linewidth}


\begin{subfigure}[t]{0.49\linewidth}
  \centering
  \includegraphics[width=\linewidth, trim=0mm 60mm 20mm 20mm, clip]{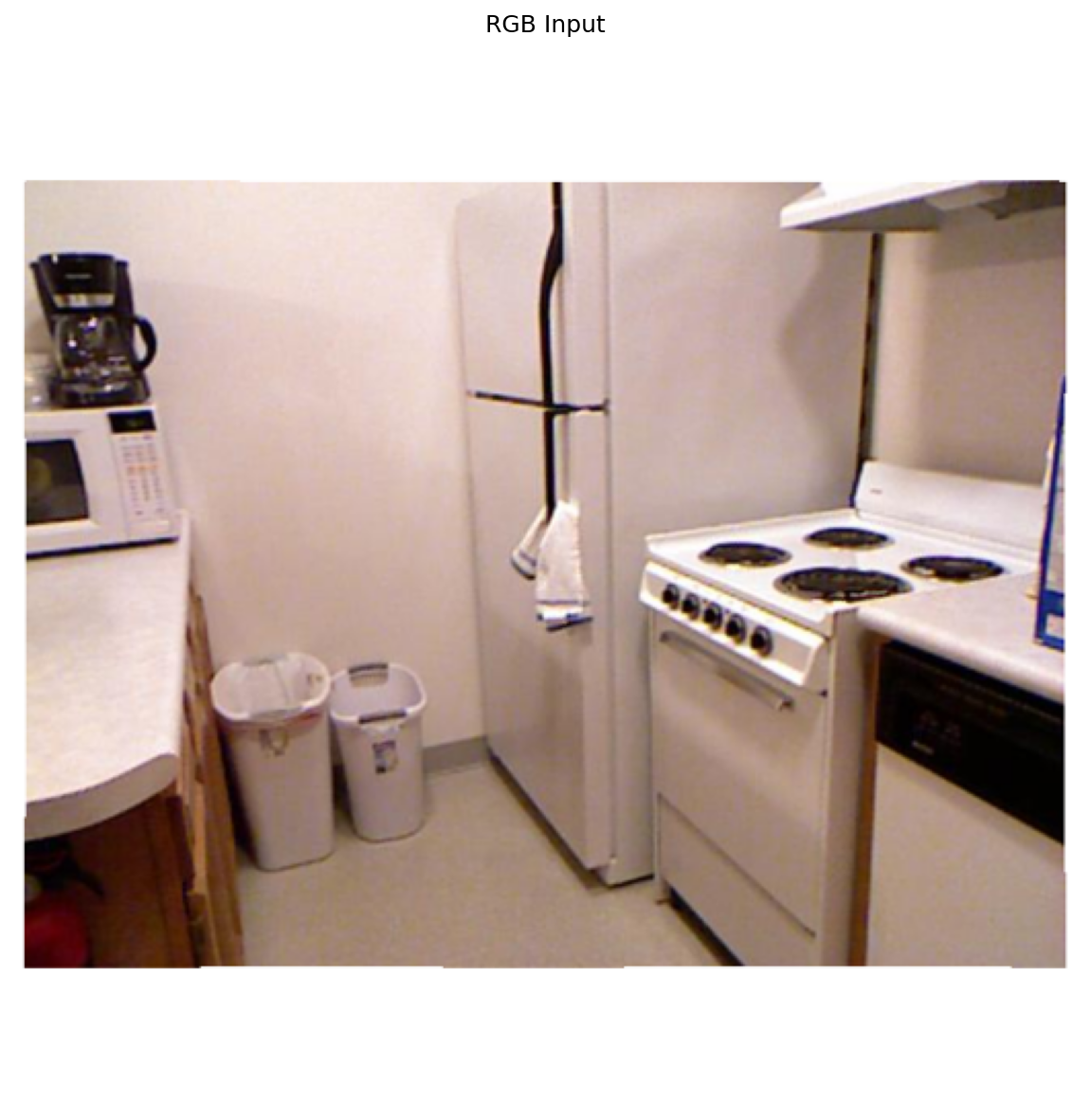}
  \caption{RGB}
\end{subfigure}
\hfill
\begin{subfigure}[t]{0.49\linewidth}
  \centering
  \includegraphics[width=\linewidth, trim=0mm 60mm 20mm 70mm, clip]{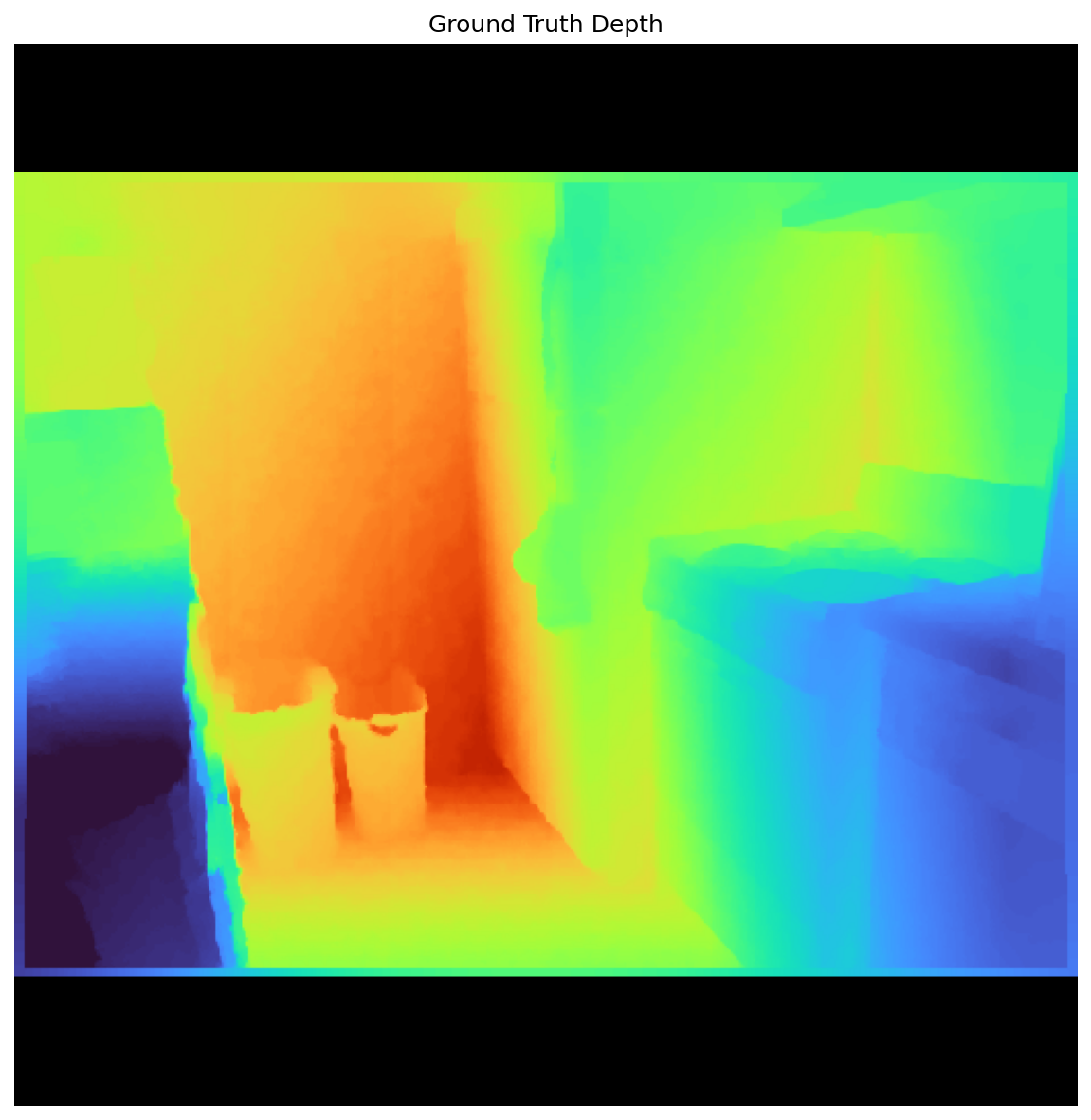}
  \caption{GT depth}
\end{subfigure}

\end{minipage}

\vspace{0mm}

\noindent\centering
\begin{subfigure}[t]{0.19\textwidth}
  \centering
  \includegraphics[width=\linewidth, trim=0mm 0mm 0mm 15mm, clip]{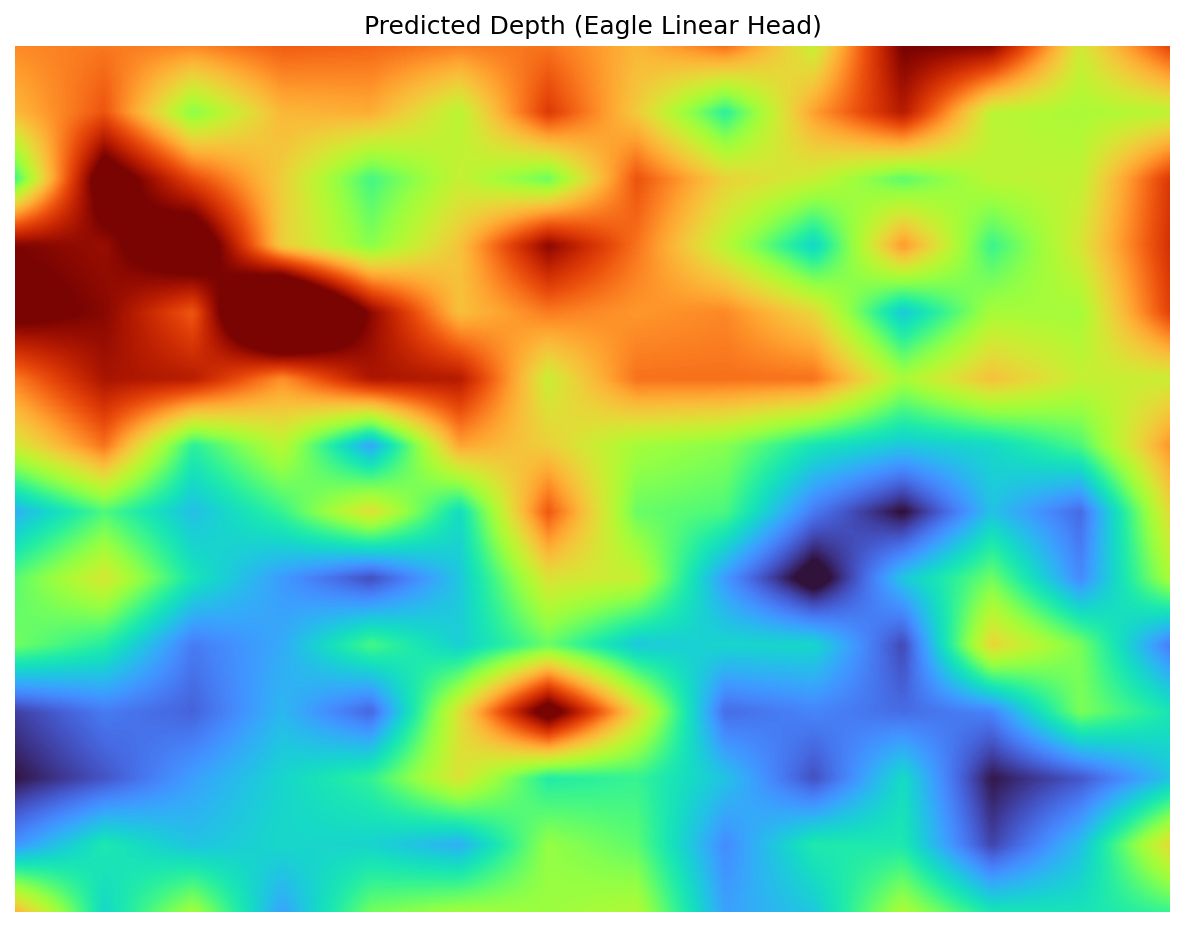}
  \caption{VLA 1}
\end{subfigure}
\hfill
\begin{subfigure}[t]{0.19\textwidth}
  \centering
  \includegraphics[width=\linewidth, trim=0mm 0mm 0mm 15mm, clip]{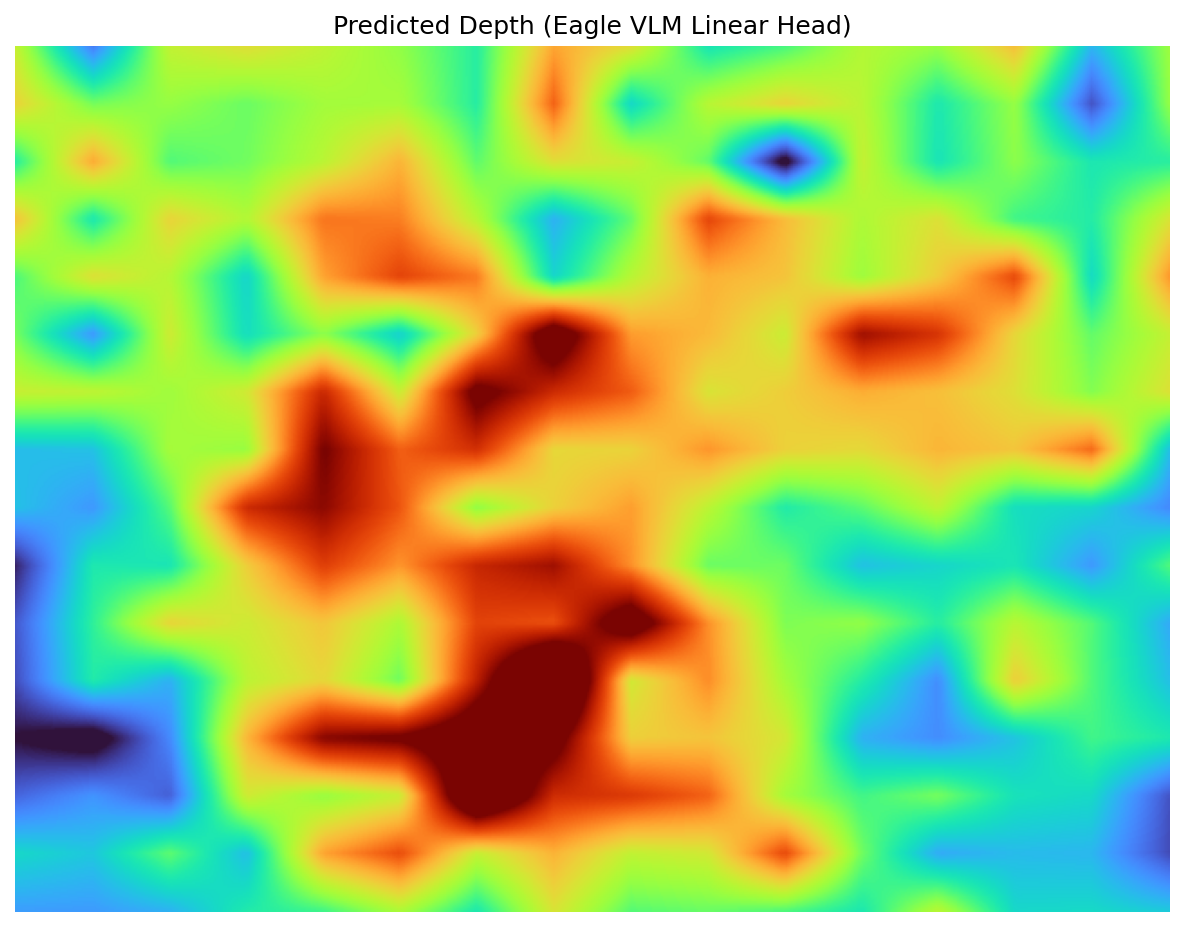}
  \caption{VLA 2}
\end{subfigure}
\hfill
\begin{subfigure}[t]{0.19\textwidth}
  \centering
  \includegraphics[width=\linewidth, trim=0mm 0mm 0mm 15mm, clip]{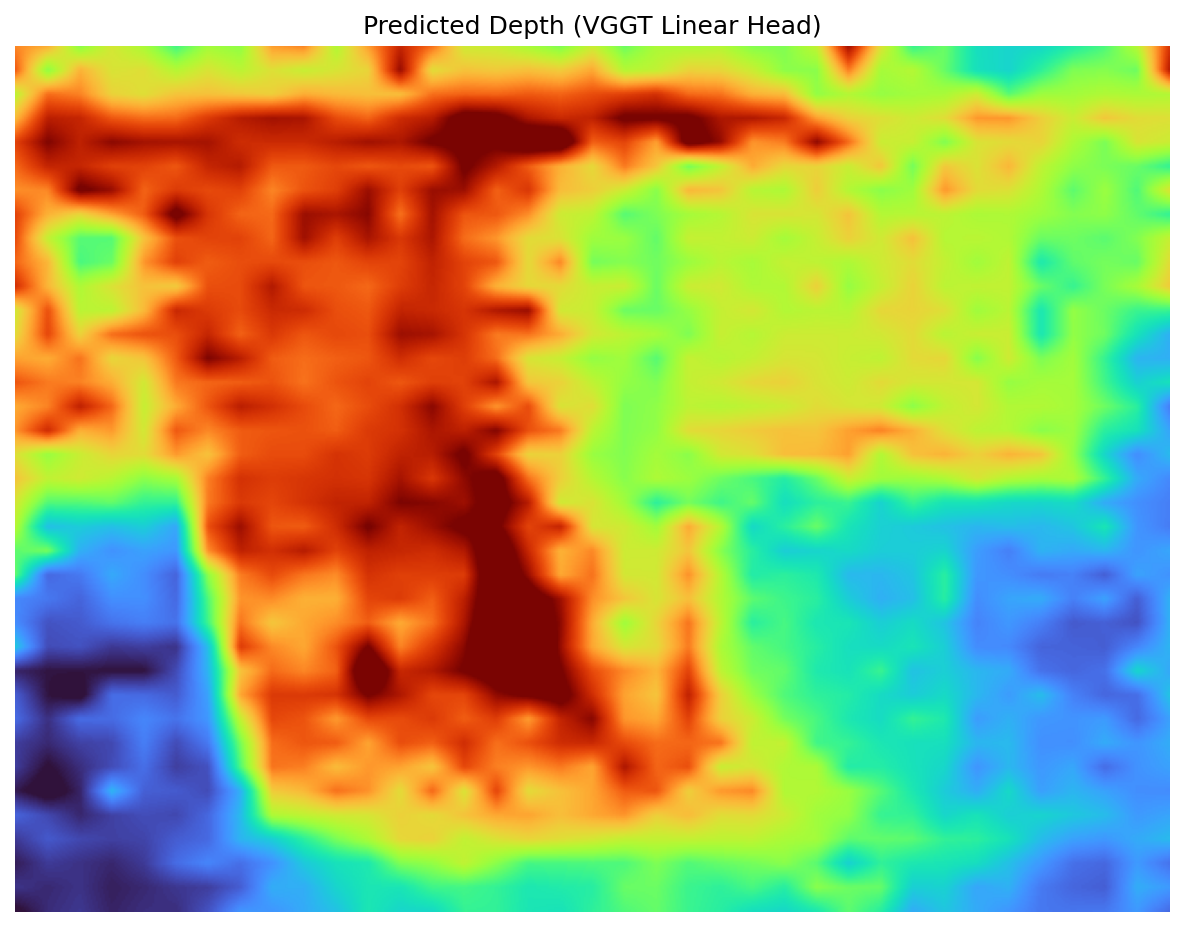}
  \caption{VGGT}
\end{subfigure}
\hfill
\begin{subfigure}[t]{0.19\textwidth}
  \centering
  \includegraphics[width=\linewidth, trim=0mm 0mm 0mm 20mm, clip]{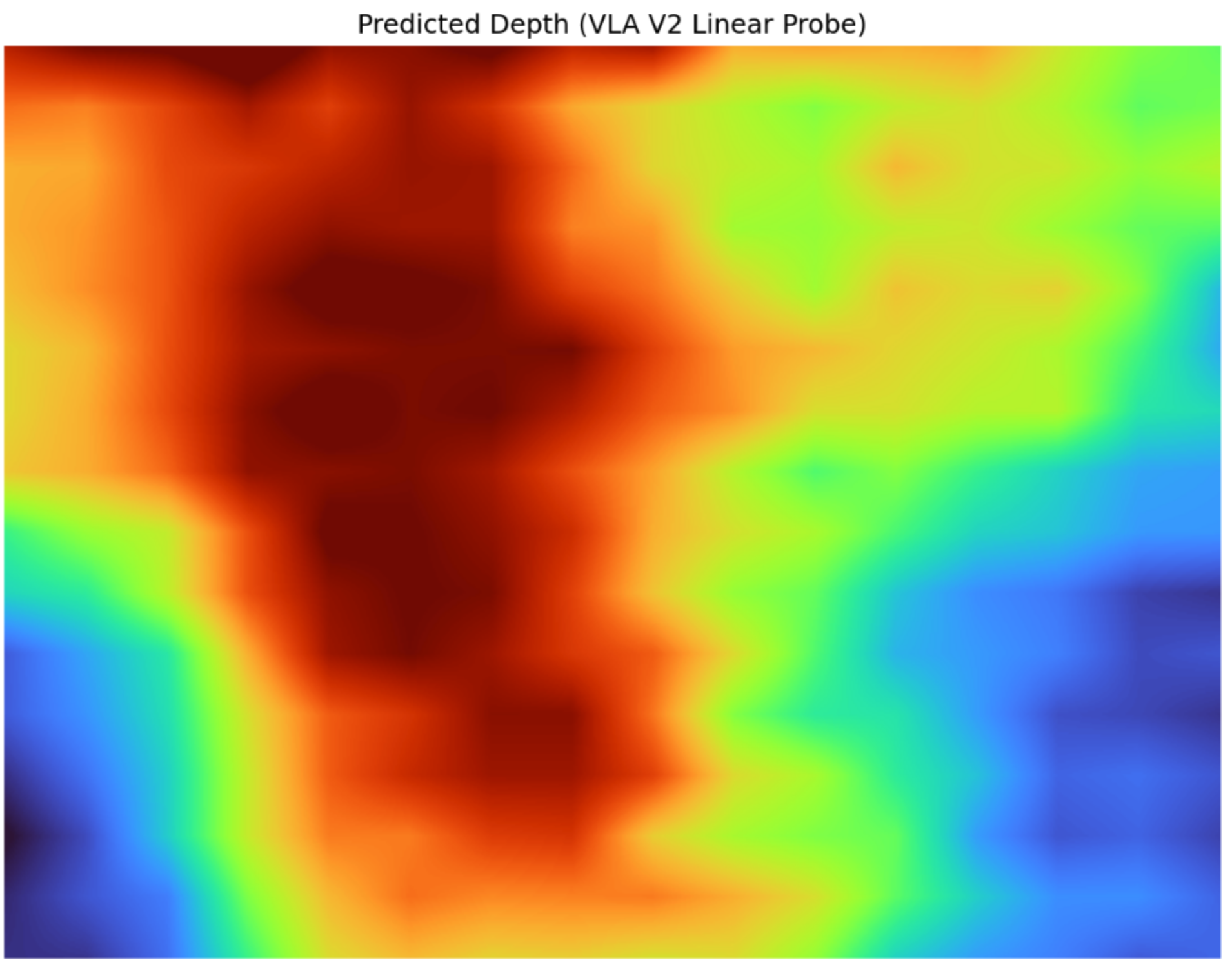}
  \caption{Early}
\end{subfigure}
\hfill
\begin{subfigure}[t]{0.19\textwidth}
  \centering
  \includegraphics[width=\linewidth, trim=0mm 0mm 0mm 15mm, clip]{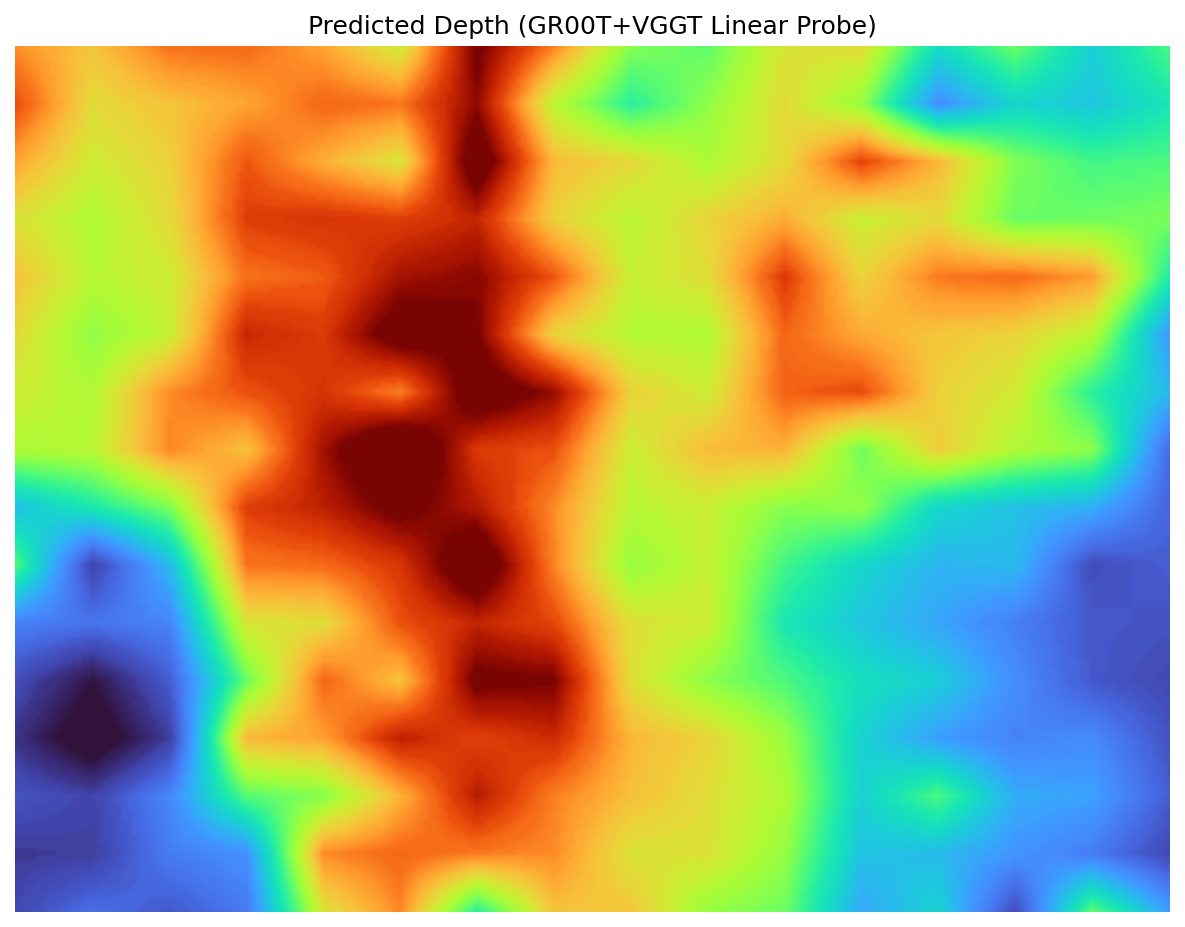}
  \caption{Late}
\end{subfigure}

\vspace{-2mm}
\captionof{figure}{Sample linear probing results: (a) RBG input, (b) ground truth (GT) depth, (c) depth predicted by the \groot vision encoder probe, (d) \groot VLM probe, (e) VGGT probe, (f) Early Fusion probe, and (g) Late Fusion probe. The poor performance of the VLA probes support the intuition that VLAs lose geometric understanding, while \emph{geometric} VLAs recover most of it thanks to the VGGT tokens.}
\label{fig:qualitative_probe_comparison}
\end{figure*}

\myParagraph{Probing Results} 
With reference to Fig.~\ref{fig:architectures}(a), we start by probing the output of the vision encoder, as well as the output of the overall GR00T VLM. Again, in both cases, the VLA is frozen at the original pre-trained weights and the linear probe is the only module that is trained.
The first two rows in Table~\ref{tab:probe_comparison} report the corresponding scores. 
For reference, we also perform linear probing using the output of the VGGT backbone and report the results in the third row of Table~\ref{tab:probe_comparison}: intuitively the VGGT tokens definitely encode geometric understanding and we can think about the gap between the performance of the VGGT probe and the VLA probes as a measure of the ``geometric gap'', which quantifies the amount of geometric information loss in the VLA.
From Table~\ref{tab:probe_comparison}, we observe that the depth understanding in VGGT is substantially higher compared to the VLA, with the RMSE almost doubling between VGGT and the VLA. The table also shows that the depth information is already lost after the vision encoder.\footnote{The attentive reader might find surprising that the probe at the output of the VLM has more depth information than the output of the vision encoder (intuitively, if the depth information is lost after the vision encoder, how can it be recovered at the output of the VLM?). This is an artifact of the analysis: the hidden dimension for vision encoder probe is 1024, while for the VLM and VGGT probes is 2048; 
therefore, the vision encoder probe is impacted by the lower capacity of the MLP.
} 

The results so far confirm that using VGGT has the potential to supplement the VLA with a better depth understanding. To further reinforce this possibility, we perform linear probing on the Early Fusion and Late Fusion models, to show that those architectures are indeed able to inject geometric information into the VLA. 
In this case, since we do not have pre-trained weights for the cross-attention fusion module, we train both the fusion module and the linear probe. In both cases, we probe the output of the VLM, to make sure geometric information is retained. Note that Spatial forcing does not entail architectural modifications hence it is excluded from this comparison.
The results are reported in the last two rows of Table~\ref{tab:probe_comparison}.
The Late Fusion architecture is able to retain geometric information. This is somewhat expected, since the VGGT tokens are injected near the probe: intuitively, the attention gate in~\eqref{eq:gate} can prioritize VGGT tokens and provide relevant geometric information to the linear probe. 
More surprisingly, even the Early Fusion architecture is able to correctly inject geometric information, achieving depth performance almost on par with VGGT.
This is nontrivial since we are injecting the VGGT tokens \emph{before} the LLM and the LLM is frozen: nevertheless, 
the LLM is still able to use the VGGT tokens to produce dense depth predictions. We provide qualitative depth prediction results in Fig.~\ref{fig:qualitative_probe_comparison}. We remark that these are the results of a \emph{linear} probe, hence ---even when accurate--- they are more noisy than the ones produced by, \eg 
a DPT~\cite{Wang25arxiv-vggt}.  
Next, we are ready to assess whether the  additional geometric information in the Early and Late Fusion models is actually helpful to improve manipulation.\footnote{Appendix~\ref{app:normalProbing} provides an additional experiment where we probe surface normal prediction. Surface normals describe how a robot can interact with an object by encoding local contact geometry and feasible force directions. 
The results are consistent with the findings in this section, further confirming the lack of geometric understanding in \groot and the potential to regain it using VGGT.}

\section{Impact of Key Design Choices} 
\label{sec:factors}

In this section, we provide a rigorous evaluation to assess the impact of key design choices on the performance of the geometric VLAs in Section~\ref{sec:architectures}.
After introducing the evaluation protocol (Section~\ref{sec:protocol}), 
we investigate the impact of the fusion strategy (Section~\ref{sec:exp-fusion}), the training data size (Section~\ref{sec:exp-midtraining}), the number of cameras (Section~\ref{sec:exp-numCam}), and the VGGT reconstruction quality (Section~\ref{sec:exp-vggt}).

\subsection{Experimental Protocol} 
\label{sec:protocol}

\myParagraph{Benchmarks}
We evaluate performance on two simulated benchmarks (RoboCasa and LIBERO) and a real benchmark (using a Unitree G1 humanoid).

\emph{RoboCasa}~\cite{Nasiriany24rss-robocasa} is a large-scale simulation benchmark. 
The benchmark includes 8 tasks: PnPCabToCounter, PnPCounterToCab, PnPCounterToMicrowave, PnPCounterToSink, PnPCounterToStove, PnPMicrowaveToCounter, PnPSinkToCounter, PnPStoveToCounter. For instance, PnPCabToCounter indicates a Pick-and-Place (PnP) task that moves an object from the cabinet to the kitchen counter. For each task, the benchmark provides 5 different episodes, each one corresponding to a different kitchen scenario and a different object to grasp. For each of the 5 episodes, we repeated 15 trials, where the appearance of the scene is randomized, for a total of 600 manipulation experiments. 
We follow the standard evaluation protocol and use three cameras (left, right, and wrist cameras).

{\small\emph{LIBERO}}~\cite{Liu23arxiv-libero} is a standard simulation benchmark and includes four tasks: {\small LIBERO-SPATIAL},
{\small LIBERO-OBJECT}, {\small LIBERO-GOAL}, and {\small LIBERO-100}. 
{\small LIBERO-100} includes 90 short-horizon tasks ({\small LIBERO-90}) and 10
long-horizon tasks ({\small LIBERO-LONG}). Each task is evaluated over 500 trials.
We follow the standard evaluation protocol, including the use of both primary and wrist-mounted cameras. 

We also benchmark on real robot data using a \emph{Unitree G1 humanoid}.
The G1 benchmark focuses on evaluating pick-and-place performance for 3 classes of objects: bottle, ball, and box. 
We repeated 90 tests across the three objects, randomizing object location, object orientation, and object instances within each class (\eg from a small wooden box to a larger cereal box for the ``box'' class). 
In each test, there is a single object on the table and the text prompt first commands the robot to pick up the object, and then to drop if off on the table.
To position the target objects in a repeatable manner across experiments, we overlaid a grid on the table surface and randomly generated the object's grid coordinates and orientation. We use the same configuration across all models.


\myParagraph{Performance Metrics} We use success rate as key performance metric, but in certain experiments we also break down the success rate across different grasping stages. For each model, we evaluate the models at the following checkpoints/epochs: 1, 5, 10, 15, 18, 20, 25, 30, 40, 50, 60, 70, 80, 90, 100, and report the best success rate across checkpoints. For each result, we compute significance levels using the two-sided McNemar's test (more details in Appendix~\ref{sec:app_mcNemar}): we observed that the randomness in the action expert can produce fluctuations in the results (even after fixing the seed for the scenario generation), hence distinguishing randomness from
actual performance differences is key (Appendix~\ref{sec:app_randomness}).

\myParagraph{Experimental Considerations}
We fix the random seeds for repeatability, without any attempt at tuning the seed for performance. 
We also fix the noise generation in the diffusion transformer of \groot.
Experimental settings are identical across all models.
We train all models using an NVIDIA A100 GPU cluster with 320 GB GPU Memory for 100 epochs. Finetuning takes 2-3 days (depending on the model), while mid-training (Section~\ref{sec:exp-midtraining}) takes around 1 week.

\myParagraph{Implementation Details}
When finetuning \groot, we follow the standard protocol and only train the action expert starting from pre-trained weights. 
For the Early Fusion and the Late Fusion model, we train the fusion module and the action expert and keek everything else frozen. 
For Spatial Forcing, we apply the alignment loss to layer 9 (out of 13) of the \groot LLM and only finetune the linear projection between the vision encoder and the LLM.
We also attempted to finetune the LLM in Spatial Forcing, but obtained worse results (Appendix~\ref{sec:extra_experiments}).

\subsection{Impact of Fusion Strategy} 
\label{sec:exp-fusion}

\begin{takeawaybox}
\textbf{Key Takeaway.}
Task-level finetuning of geometric VLAs combining \groot-N1.5 and VGGT does not lead to a statistically significant increase in the success rate.
The Early Fusion geometric VLA appears to be more promising in real world experiments.
\end{takeawaybox}

\noindent
In this section, we compare the three geometric VLA models from Section~\ref{sec:architectures}.

\myParagraph{RoboCasa Results}
Table~\ref{tab:robocasaResultsTransposed} reports success rates across the 8 RoboCasa tasks, including both the implemented geometric VLAs (Early Fusion, Late Fusion, and Spatial Forcing) and baselines from the literature. For each, we also report $p$ values for statistical significance, all computed with respect to the \groot-N1.5 baseline. For each geometric VLA, we finetune on the specific task using the demonstrations provided by RoboCasa (around 300 demonstrations per task). Moreover, we finetune and test \groot-N1.5 to ensure a fair comparison. 
We postpone the discussion of the last row of the table to Section~\ref{sec:exp-midtraining}.

\begin{table*}[ht]
\centering
\scriptsize
\resizebox{\textwidth}{!}{%
\begin{tabular}{l|llllllll|l}
\hline
& \textbf{CabToCtr}
& \textbf{CtrToCab}
& \textbf{CtrToMicrowave}
& \textbf{CtrToSink}
& \textbf{CtrToStove}
& \textbf{MicrowaveToCtr}
& \textbf{SinkToCtr}
& \textbf{StoveToCtr} 
& \textbf{Average} \\
\hline

DP3\cite{ze20243d}
& 4.0 & 2.0 & 6.0 & 0.0 & 0.0 & 6.0 & 0.0 & 0.0 & 2.3\\

Pi0\cite{black2024pi0}
& 28.0 & 18.0 & 36.0 & 70.0 & 36.0 & 22.0 & 16.0 & 44.0 & 33.8\\

Pi0-Fast\cite{pertsch2025fast}
& 30.0 & 48.0 & 20.0 & 56.0 & 64.0 & 46.0 & 62.0 & 60.0 & 48.3 \\

RS-CL\cite{kim2025contrastive}
& \cellcolor{best}\textbf{60.0}
& 68.0
& 40.0
& 68.0
& 72.0
& 48.0
& 68.0
& 54.0
& 59.0 \\

DP-VLA\cite{han2024dual}
& 10.0 & 32.0 & 56.0 & 30.0 & 22.0 & 18.0 & 56.0 & 62.0 & 35.8\\

GR00T N1 \cite{bjorck2025gr00t}
& 20.0 & 36.0 & 13.0 & 10.0 & 24.0 & 16.0 & 33.0 & 29.0 & 22.6\\

Video Policy\cite{hu2024video}
& 48.0 & 52.0 & 22.0 & 48.0 & 54.0 & 28.0 & 56.0 & 70.0 & 47.3\\

\hline

\groot-N1.5 
& 42.7
& \cellcolor{best}\textbf{74.7}
& \cellcolor{second}73.3
& \cellcolor{second}93.3
& 77.3
& 58.7
& 65.3
& \cellcolor{second}88.0
& \cellcolor{second}71.7\\

Early Fusion
& 32.0\,{\scriptsize (p=0.186)}
& 69.3\,{\scriptsize (p=0.289)}
& 65.3\,{\scriptsize (p=0.377)}
& 88.0\,{\scriptsize (p=0.388)}
& 73.3\,{\scriptsize (p=0.664)}
& 62.7\,{\scriptsize (p=0.742)}
& 80.0\,{\scriptsize (p=0.019)}
& 86.7\,{\scriptsize (p=1.000)}
& 69.7\,{\scriptsize (p=0.399)} \\

Late Fusion 
& 46.7\,{\scriptsize (p=0.710)}
& \cellcolor{second}72.0\,{\scriptsize (p=0.625)}
& \cellcolor{best}\textbf{74.7}\,{\scriptsize (p=1.000)}
& 85.3\,{\scriptsize (p=0.146)}
& 69.3\,{\scriptsize (p=0.307)}
& \cellcolor{best}\textbf{69.3}\,{\scriptsize (p=0.115)}
& 69.3\,{\scriptsize (p=0.689)}
& 81.3\,{\scriptsize (p=0.267)}
& 71.0\,{\scriptsize (p=0.806)} \\


Spatial Forcing
& 29.3\,{\scriptsize (p=0.123)}
& 68.0\,{\scriptsize (p=0.227)}
& 66.7\,{\scriptsize (p=0.499)}
& 76.0\,{\scriptsize (p<0.001)}
& 72.0\,{\scriptsize (p=0.454)}
& 60.0\,{\scriptsize (p=1.000)}
& \cellcolor{best}\textbf{84}\,{\scriptsize (p=0.007)}
& \cellcolor{best}\textbf{90.7}\,{\scriptsize (p=0.804)}
& 68.3\,{\scriptsize (p=0.154)} \\

\hline
\hline

Early Fusion (mid-trained)
& \cellcolor{second}52.0\,{\scriptsize (p=0.281)}
& \cellcolor{second}72.0\,{\scriptsize (p=0.727)}
& 69.3\,{\scriptsize (p=0.700)}
& \cellcolor{best}\textbf{94.7}\,{\scriptsize (p=1.000)}
& \cellcolor{best}\textbf{80.0}\,{\scriptsize (p=0.815)}
& \cellcolor{second}68.0\,{\scriptsize (p=0.189)}
& \cellcolor{second}81.3\,{\scriptsize (p=0.023)}
& 84.0\,{\scriptsize (p=0.607)}
& \cellcolor{best}\textbf{75.2}\,{\scriptsize (p=0.104)} \\

\hline
\end{tabular}}
\caption{VLA performance comparison on the RoboCasa benchmark. Green indicates best result per column; yellow indicates second best (ties share the same color). $p$-values are computed against the \groot-N1.5 baseline. Results for the first 7 rows are borrowed from the corresponding paper.}
\label{tab:robocasaResultsTransposed}
\end{table*}

From a quick glance at the table, the results show that in average the performance of the geometric VLAs is subpar compared to the \groot baseline (\eg 69.7\% for Early Fusion vs. 71.2\% for the baseline). However, closer inspection of the $p$ values suggests that the baseline results are just not significantly different from the baseline: 
a $p$ value larger than $p=0.05-0.1$ indicates the differences are not statistically significant and might be the result of randomness. 
We remark that each $p$ value in the ``Average'' column is not the average of the $p$ values of that row, but it is recomputed accounting for the success rates of the runs across all tasks. 
We draw similar conclusions from testing on the LIBERO benchmark, see Appendix~\ref{sec:extra_experiments} for further details.
In conclusion, basic finetuning of geometric VLAs does not fundamentally change success rate in simulated benchmarks. 
As we will show in the next sections, the conclusion drastically changes 
when increasing the amount of training data or the sensor configuration.
\begin{table*}[ht]
\centering
\resizebox{\textwidth}{!}{%
\begin{tabular}{l|ccccc}
\hline
& \textbf{Approach} 
& \textbf{Grasp} 
& \textbf{Lift} 
& \textbf{Placement} 
& \textbf{Overall} \\
\hline

\groot-N1.5
& \cellcolor{second}57.78 & 51.92 & 85.19 & \cellcolor{best}\textbf{86.96} & 22.22 \\

Early Fusion
& \cellcolor{best}\textbf{84.44}\,{\scriptsize (p<0.001)} & \cellcolor{best}\textbf{60.53}\,{\scriptsize (p=0.824)} & \cellcolor{second}89.13\,{\scriptsize (p=1.000)} & 65.85\,{\scriptsize (p=0.180)} & \cellcolor{best}\textbf{27.78}\,{\scriptsize (p=0.511)} \\

Late Fusion 
& 57.78\,{\scriptsize (p=0.855)} & \cellcolor{second}59.62\,{\scriptsize (p=1.000)} & \cellcolor{best}\textbf{93.55}\,{\scriptsize (p=1.000)} & \cellcolor{second}79.31\,{\scriptsize (p=0.625)} & \cellcolor{second}25.56\,{\scriptsize (p=0.710)} \\

\hline
\end{tabular}
}
\caption{VLA performance comparison on (real) Unitree G1 benchmark. The table reports the overall success rate, as well as the breakdown of the success rate in terms of 
percentage of cases where the robot successfully approached, grasped, lifted, and placed the target object. Green indicates best result per column; yellow indicates second best. 
}
\label{tab:unitreeResultsBreakdown}
\end{table*} 

\myParagraph{Unitree G1 Results}
Table~\ref{tab:unitreeResultsBreakdown} reports the results of the real manipulation benchmark on the Unitree G1.
Since each experiment is time consuming, we focus on the Early Fusion and the Late Fusion models. 
In this case, we also provide a breakdown of the performance across grasping stages: the Approach stage is successful when the robot 
approaches and touches the object of interest; the Grasp stage is successful when the robot is able to grasp the object; 
the Lift stage is successful when the object is correctly lifted above the table; the Placement stage is successfully when the object 
is correctly placed back on the table. If all the stages are successful, the overall episode is considered successful.
Several considerations are in order. First of all, the overall success rate is higher than baseline ($22.22\%$) for both the Early Fusion ($27.78\%$) and Late Fusion model ($25.56\%$). 
However, the results again lack strong statistical significance, with $p$ values above 0.1. 
At the same time, the success rate of the approach stage is statistically better than baseline for the Early Fusion model, 
achieving $84.44\%$ success rate compared to $57.78\%$ of the \groot baseline. 
This might indicate that Early Fusion is able to leverage geometric information to better identify the target object and more precisely move the gripper towards it.
In average, Grasp and Lift performance is better for the Early and Late Fusion models compared to baseline, while the placement is slightly worse, but again 
these observations are assigned relatively large $p$ values.
In practice, we observed the Early Fusion approach to more reliably grasp small objects (\eg the small ball and the small box in our benchmark), which was very challenging for the \groot baseline instead. 
The performance gain of the Late Fusion approach is more modest. We conjecture that the VLM (and, in particular, the LLM within the LLM) is more flexible in processing new information (as the one provided by the Early Fusion) ---an observation confirmed by the plasticity observed in the probing experiment of Section~\ref{sec:probing};
on the other hand, the action expert is relatively ``rigid'' and unable to make good use of additional data sources (as in the Late Fusion).
Next, we go deeper into our investigation and focus on other aspects that make the advantage of using \GFMs more pronounced. 
We mostly focus on the Early Fusion model, mostly due to the fact that it stands out in real experiments, and we postpone details about other models to 
Appendix~\ref{sec:extra_experiments}.


\subsection{Impact of Training Data Scaling}
\label{sec:exp-midtraining}

\begin{takeawaybox}
\textbf{Key Takeaway.}
Early Fusion's performance largely improves and becomes better than baseline (in a statistically significant manner) 
after training it on a larger amount of data before finetuning on the specific task.
\end{takeawaybox}

This section shows that we can draw stronger conclusions about the impact of geometric information if we scale up the training data.
For this experiment, we first trained the Early Fusion model on the entire RoboCasa dataset (all 8 tasks) starting from the \groot's pretrained weights.
The basic idea is that this ``mid-training'' allows the network to better adjust to the VGGT tokens and make better use of the resulting geometric information.
After mid-training, we finetune on a specific RoboCasa task, as done in the previous section. 
The corresponding results are reported in the last row of Table~\ref{tab:robocasaResultsTransposed}.
Interestingly, now the Early Fusion approach with mid-training becomes substantially better than baseline, with $p$ values indicating increased statistical significance. 
To further demonstrate that the advantage is due to VGGT rather than other potential information leakage induced by the mid-training, 
 Appendix~\ref{sec:extra_experiments} shows that Early Fusion outperforms \groot even when both are mid-trained using the same protocol.
This observation is important because it suggests that scaling up the training of geometric VLAs may unlock their full potential.
Such a large-scale training is indeed possible since it would not require additional training data.

\subsection{Impact of the Number of Cameras}
\label{sec:exp-numCam}

\begin{takeawaybox}
\textbf{Key Takeaway.}
Even without mid-training, Early Fusion's performance becomes statistically better than baseline when using a single-camera setup.
\end{takeawaybox}

While we have observed that mid-training already makes the Early Fusion model statistically better than baseline, one might also ponder if ---without mid-training--- 
there are factors that could make VGGT more useful for manipulation.
One may conjecture that \groot might already be able to infer depth information from multiple camera views. 
On the other hand, the advantage of using a geometric foundation model might be more pronounced when using a single camera, since in that case the VLA would not be able to obtain dense geometric understanding via multi-view geometry.
To test this hypothesis, we finetune both the Early Fusion model and the \groot baseline using a single camera (in particular, the left camera) in Robocasa.
Table~\ref{tab:robocasaResultsSingleCamera} reports success rates for both single-camera models. 
For both, as expected, the overall performance drastically drops, confirming that multi-view inputs are key to successful grasping.
More interestingly, the advantage of Early Fusion becomes more pronounced ($21.5\%$ vs. $17.2\%$ for the baseline) and statistically significant ($p<0.05$), confirming our intuition.  
This might suggest that, in sensor-constrained settings, geometric foundation models provide a viable option to boost performance.

\begin{table*}[t!]
\centering
\scriptsize
\resizebox{\textwidth}{!}{%
\begin{tabular}{l|llllllll|l}
\hline
\textbf{Method} 
& \textbf{CabToCtr}
& \textbf{CtrToCab}
& \textbf{CtrToMicrowave}
& \textbf{CtrToSink}
& \textbf{CtrToStove}
& \textbf{MicrowaveToCtr}
& \textbf{SinkToCtr}
& \textbf{StoveToCtr} 
& \textbf{Average} \\
\hline

\groot-N1.5 
& \cellcolor{best}\textbf{12.0}
& 8.0
& 33.3
& \cellcolor{best}\textbf{28.0}
& 13.3
& 12.0
& 26.7
& 4.0
& 17.2 \\

Early Fusion 
& 10.7\,{\scriptsize (p=1.000)}
& \cellcolor{best}\textbf{9.3}\,{\scriptsize (p=1.000)}
& \cellcolor{best}\textbf{38.7}\,{\scriptsize (p=0.503)}
& 17.3\,{\scriptsize (p=0.115)}
& \cellcolor{best}\textbf{29.3}\,{\scriptsize (p=0.008)}
& \cellcolor{best}\textbf{18.7}\,{\scriptsize (p=0.302)}
& \cellcolor{best}\textbf{34.7}\,{\scriptsize (p=0.307)}
& \cellcolor{best}\textbf{13.3}\,{\scriptsize (p=0.039)}
& \cellcolor{best}\textbf{21.5}\,{\scriptsize (p=0.030)} \\

\hline
\end{tabular}}
\caption{RoboCasa results with single-camera (no mid-training). Green indicates best result per column. $p$ values are computed against \groot-N1.5 (single camera).}
\label{tab:robocasaResultsSingleCamera}
\end{table*}

\subsection{Impact of VGGT Performance}
\label{sec:exp-vggt}

\begin{takeawaybox}
\textbf{Key Takeaway.}
While the Early Fusion model's success rate is largely task dependent, 
the success rate is also correlated with VGGT's performance (\ie smaller depth errors in VGGT correlate with 
highest success rates).
\end{takeawaybox}

\begin{figure*}[h!]
\centering

\begin{subfigure}[t]{0.62\textwidth}
\vspace{0pt}
    \centering
    \begin{overpic}[width=\linewidth, trim=0mm 0mm 0mm 16mm, clip]{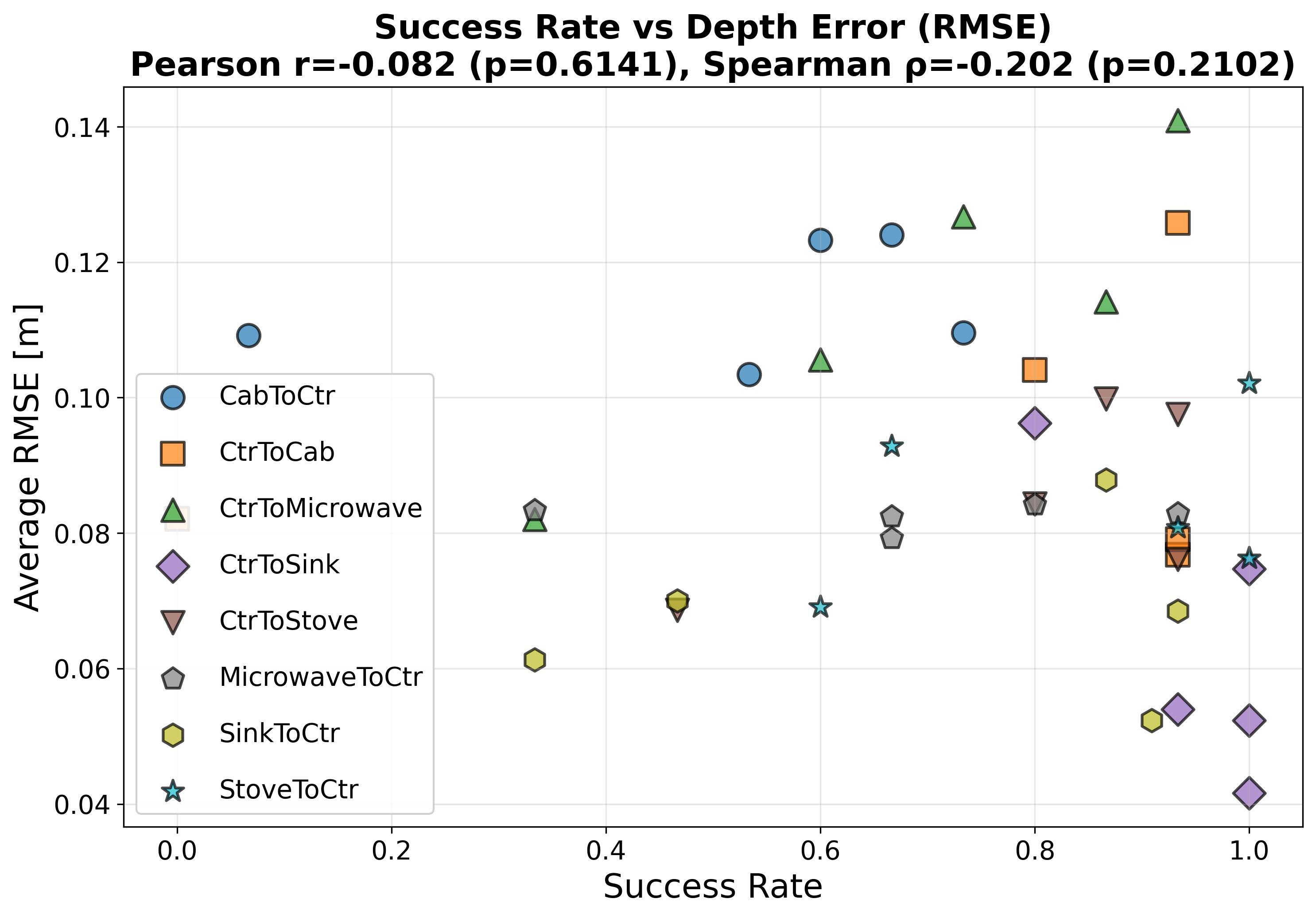}
    \put(35,0){(a)}
    \end{overpic}
\end{subfigure}
\hfill
\begin{subfigure}[t]{0.36\textwidth}
\vspace{0pt}
\centering

\begin{tabular}{cccc}
& {Left} & {Wrist} & {Right}\\
 
\rotatebox{90}{\hspace{3mm}{RGB}} &
\includegraphics[width=0.28\textwidth]{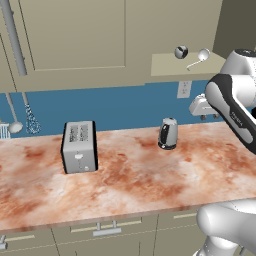} &
\includegraphics[width=0.28\textwidth]{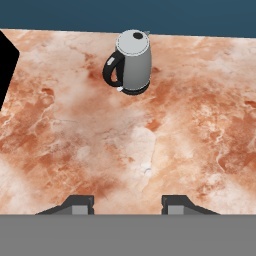} &
\includegraphics[width=0.28\textwidth]{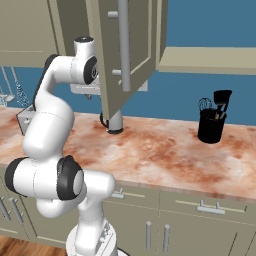} \\

\rotatebox{90}{VGGT} &
\includegraphics[width=0.28\textwidth, trim=0mm 0mm 30mm 12mm, clip]{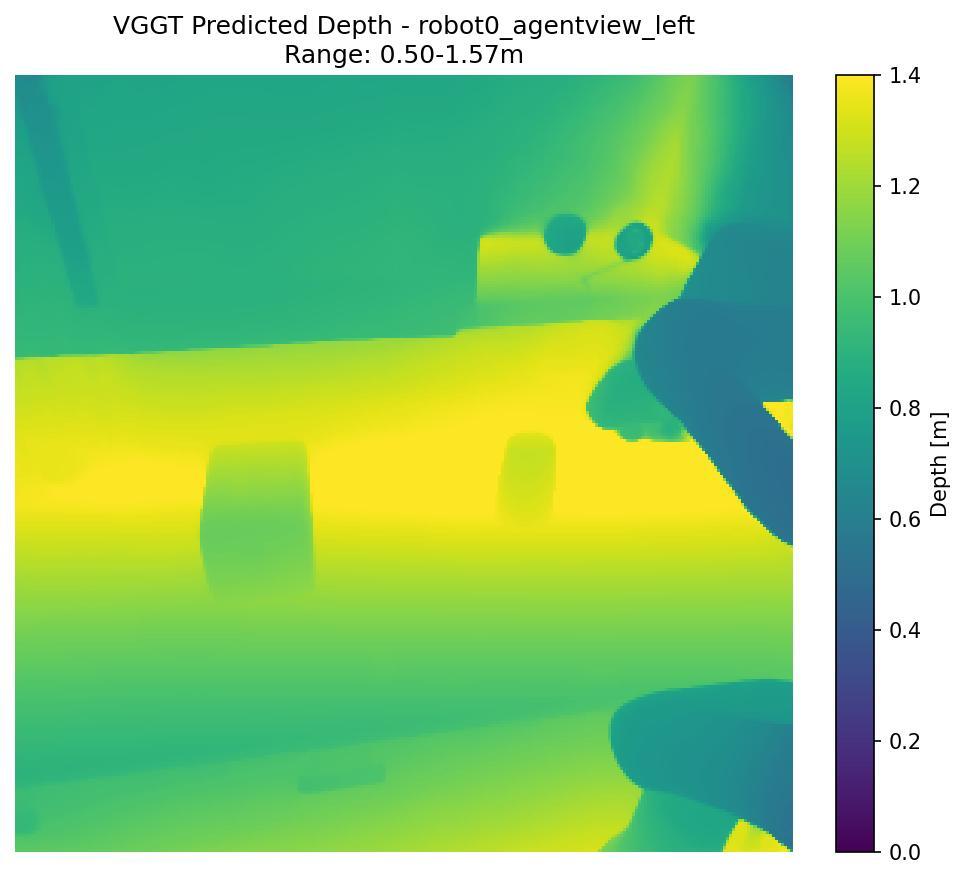} &
\includegraphics[width=0.28\textwidth, trim=0mm 0mm 30mm 12mm, clip]{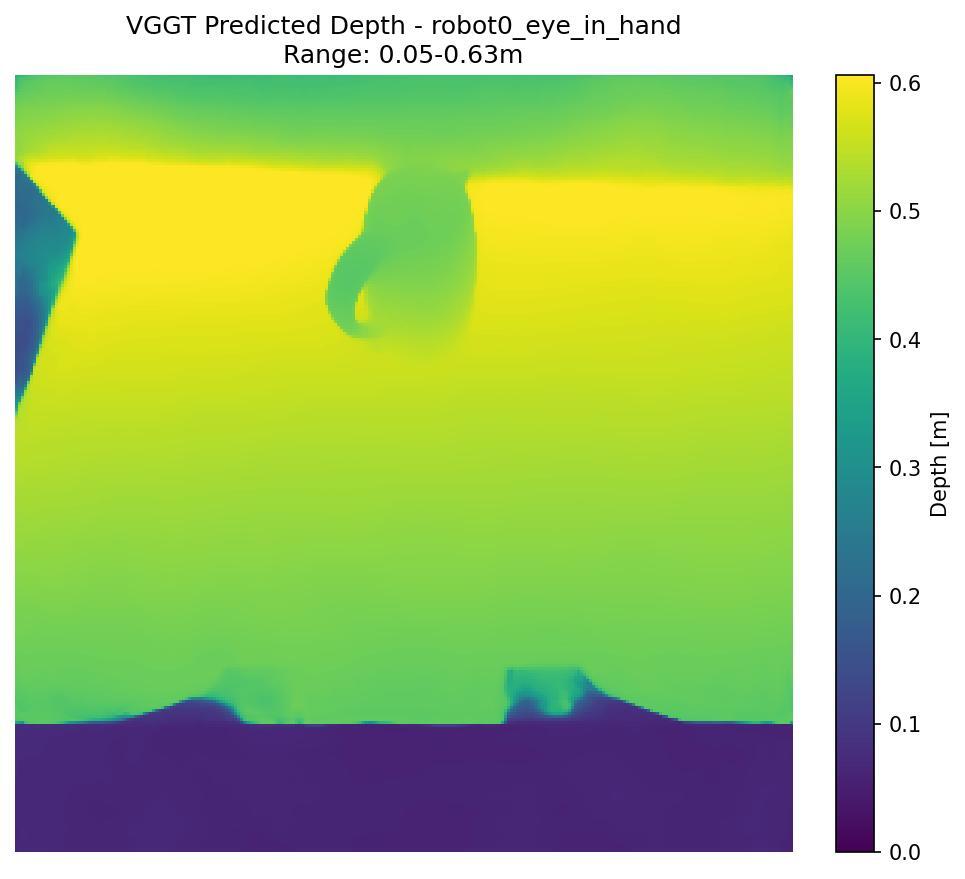} &
\includegraphics[width=0.28\textwidth, trim=0mm 0mm 30mm 12mm, clip]{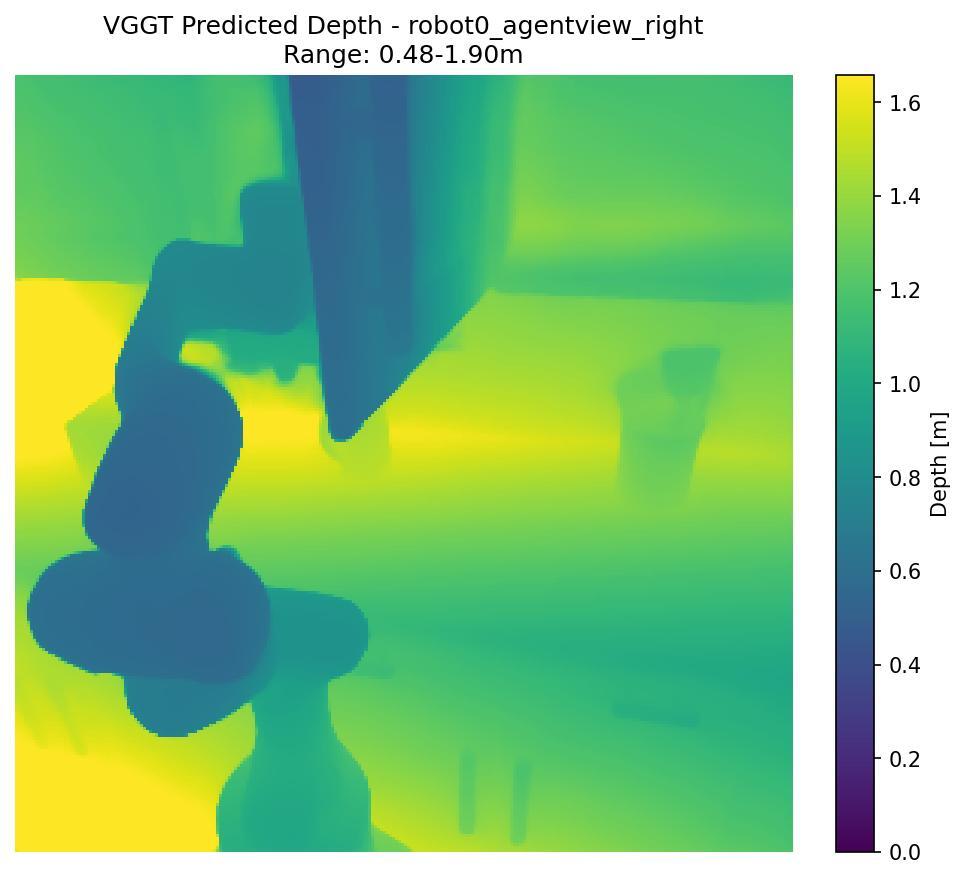} \\

\rotatebox{90}{\hspace{2mm}GT} &
\includegraphics[width=0.28\textwidth, trim=0mm 0mm 30mm 12mm, clip]{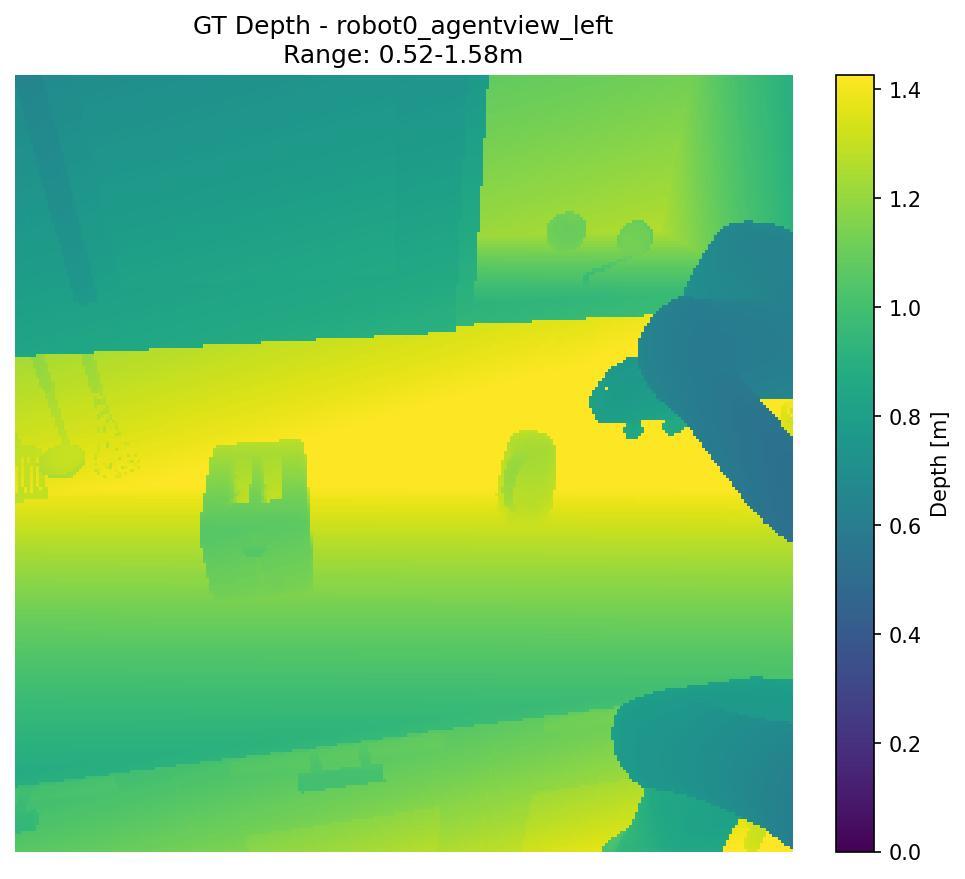} &
\includegraphics[width=0.28\textwidth, trim=0mm 0mm 30mm 12mm, clip]{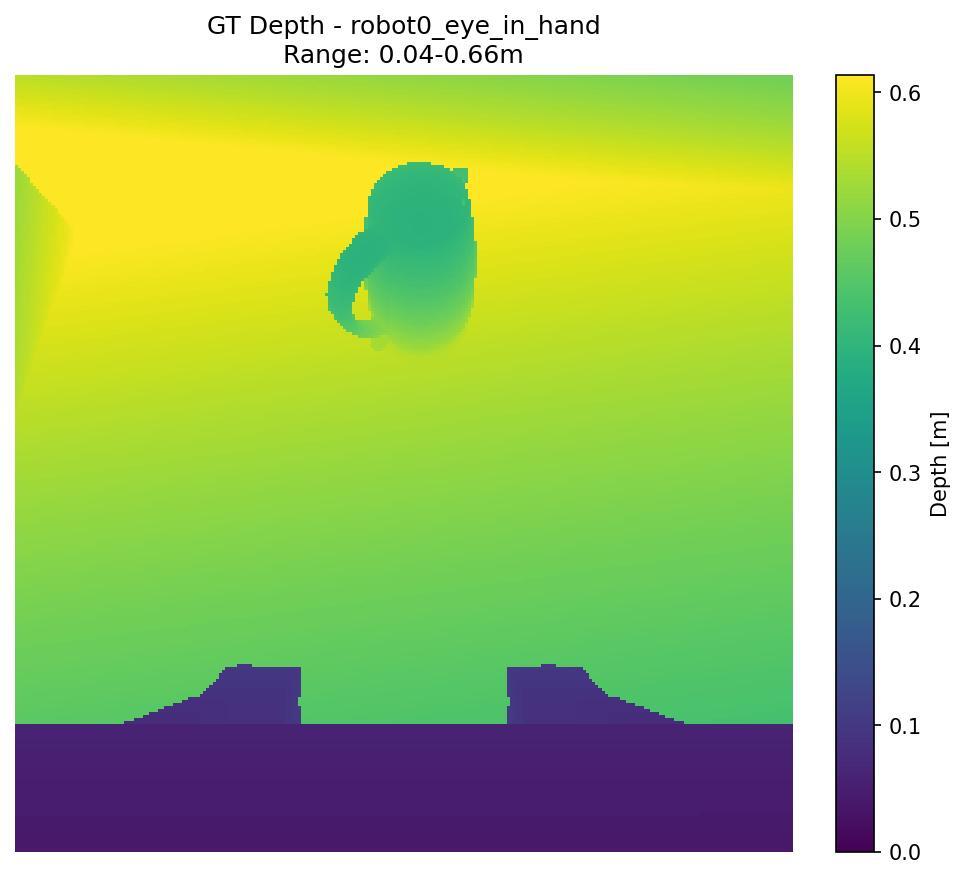} &
\includegraphics[width=0.28\textwidth, trim=0mm 0mm 30mm 12mm, clip]{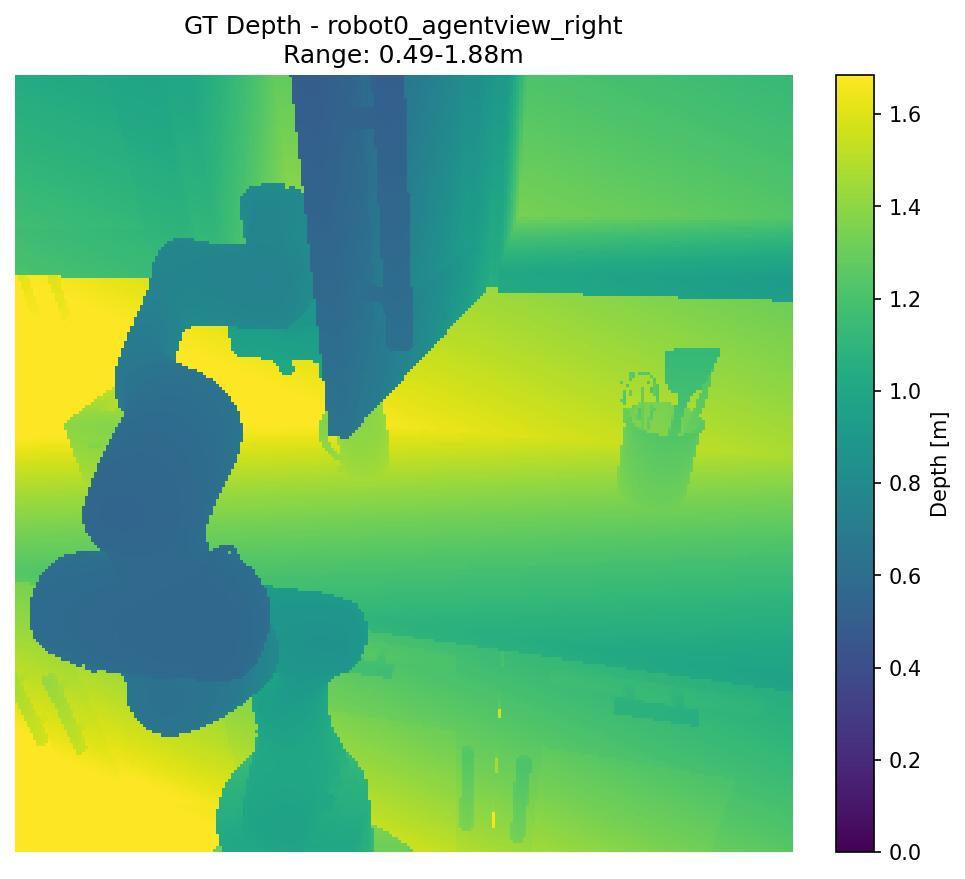} \\

&&(b)&\\
\end{tabular}


\end{subfigure}
\vspace{-2mm}
\caption{(a) VGGT depth error vs. manipulation success rate in RoboCasa. (b) Sample RGB images and predicted vs. ground-truth depth, for the left, wrist, right cameras.}
\label{fig:seven_panel_layout}
\end{figure*}

Finally, we test the impact of the VGGT reconstruction performance on the success rate of the resulting geometric VLA.
Intuitively, we expect that in cases where VGGT is out-of-distribution or exhibits larger errors, such errors would propagate to
the VLA as well. Conversely, scenarios with more accurate VGGT performance might achieve better success rates.
Fig.~\ref{fig:seven_panel_layout}(a) plots ---for each of the 5 episodes of the 8 RoboCasa tasks--- the success rate for that episode versus the average (RMSE) depth error of the VGGT reconstruction. The depth error is averaged across all images and cameras in the episode. 
First of all, Fig.~\ref{fig:seven_panel_layout}(a) shows that the depth errors are relatively small, confirming the outstanding generalization capabilities of VGGT, which performs well despite the relatively low-resolution camera renderings (qualitative results in Fig.~\ref{fig:seven_panel_layout}(b)).
More importantly, we can inspect the correlation between depth quality and success rate.
While the success rate is largely impacted by other factors beyond the VGGT reconstruction quality (such as the task complexity), we can see that the right-hand side of the plot shows a decreasing trend, meaning that lower RMSE corresponds to higher success rates. More formally, we can measure
the correlation between depth reconstruction error and success rate: using the data in Fig.~\ref{fig:seven_panel_layout}(a), we 
compute the Spearman correlation coefficient and obtain $\rho = -0.202$, which confirms a mild negative correlation between the variables.
This observation is important since it suggests that finetuning VGGT in the target environment might further top-off the performance of the geometric VLA.


\section{Limitations} 
\label{sec:limitations}
\vspace{-2mm}

Despite our best effort to provide a rigorous evaluation of the impact of geometric foundation models on VLAs, this work is not without limitations.
First of all, the results are limited to our choice of VLA (\groot-N1.5) and \GFM (VGGT). There are many other VLAs and \GFMs, and despite the presence of architectural similarities, we cannot generalize our conclusions beyond the combination we tested.
Second, most of the results are on simulation benchmarks (except our analysis on the Unitree G1). While this is common practice in the field, 
it might not fully reflect real performance (indeed we observed that models using \GFMs show stronger performance on real data).
This issue is compounded by the fact that some of the simulation benchmarks are saturated, hence limiting the margin of improvement that can be shown. 
Third, while we tried to follow design choices used in related work, we did not ablate the impact of other choices (\eg which layer to apply Spatial Forcing to,
or alternative fusion strategies not relying on cross-attention); the scope of our experimental analysis is already very broad, with some of the training experiments taking close to a week to complete, and exploring the entire design space is simply impractical.
Finally, while we believe that adding statistical significance (and avoiding tuning random seeds) is crucial to draw rigorous conclusions, 
the corresponding $p$ values are not binary, and still leave a margin for errors; we acknowledge this uncertainty in our evaluation.

\vspace{-2mm}
\section{Conclusions} 
\label{sec:conclusion}
\vspace{-2mm}

We provide an analysis of the impact of injecting geometric information via Geometric Foundation Models (\GFMs) into modern VLAs.
First, we introduced three different strategies to inject geometric knowledge in VLAs, providing a unifying lens over the recent literature.
Second, we quantified the gap in geometric understanding between a VLA (\groot-N1.5) and a \GFM (VGGT) using linear probing,
and show such a gap can be filled by architectures leveraging \GFMs. 
Third, we discussed factors impacting performance, including both architectural choices (\ie choice of fusion strategy) as well as non-architectural choices (including 
the size of the training data, the number of cameras, and the quality of the VGGT reconstruction). 
As a result, this work provides non-trivial insights into the performance of novel VLAs integrating geometric foundation models, that we 
believe can trigger new and impactful research in this area.


\clearpage

\bibliographystyle{splncs04}
\bibliography{refs.bib, refs-spark.bib}

\clearpage

\appendix
\setcounter{table}{0}
\renewcommand{\thetable}{A.\arabic{table}}


\section{Appendix: Details about Architectures} 
\label{sec:architecture_details}

This appendix complements the discussion in Section~\ref{sec:architectures}, where we introduced the Early and Late Fusion models, and discussed the 
implementation of the fusion module as a cross-attention layer with attention gating.

As we mentioned in Section~\ref{sec:architectures}, the fusion module  $\tilde{\MX} = {\tt fuse}(\MX,\MG)$ has the goal of fusing 
\groot tokens $\MX \in \Real{P_x \times D_x}$ with VGGT tokens $\MG \in \Real{P_g \times D_g}$. 
We only select the visual tokens from \groot; similarly, we only select the image tokens from VGGT (and disregard the register tokens).
Note that \groot and VGGT tokenize the image into a different number of patches and use different image pre-processing pipelines, hence usually $P_x \neq P_g$. 

\myParagraph{Positional Embeddings}
Since tokens in $\MX$ and $\MG$ are associated to image patches, we use learned positional embeddings to allow the model 
to retain the association between tokens and their spatial arrangement in the image. 
In particular, 
we add modality-specific positional embeddings:
\begin{align}
\MX \leftarrow \MX + \mathbf{E}^{2D}, \qquad 
\MG \leftarrow \MG + \mathbf{E}^{3D}
\end{align}
where $\mathbf{E}^{2D}$ and $\mathbf{E}^{3D}$ encode spatial layouts for the \groot and VGGT tokens. 

\myParagraph{Cross Attention}
After adding learnable positional embeddings, the tokens are passed through a cross-attention layer, following the computation in eqs.~\eqref{eq:QKV}-\eqref{eq:projWO}.
In our implementation, we follow standard practice and parameterize the linear projections ${\MW}_Q, {\MW}_K ,{\MW}_V, \MW_O$ as LoRA (Low-Rank Adaptation) layers~\cite{hulora} 
with rank $8$.

\myParagraph{Attention Gating} 
Not all 3D information is equally useful at every spatial location.
We therefore introduce a learnable gate $\MA$ and apply it to the cross-attention results $\MZ$:  
\begin{equation}
\tilde{\MX} =
\MX + \MA \odot \MZ
\end{equation}
The gate is applied element-wise and is designed to select relevant portions of the residual correction $\MZ$, to produce the fused tokens $\tilde{\MX}$.
We initialize the gate close to zero, in order to more gradually introduce the residual $\MZ$ from the VGGT tokens; 
indeed, we observed that the residual ---if introduced too suddenly---  could bring the action expert off-distribution and disrupt the finetuning process (see experiments without the attention gate in Appendix~\ref{sec:ablation-early-1}). 
We also experimented with making the gate input-dependent:
\begin{equation}
\MA =
\sigma\left(
\MX \MG_a + \vb_a
\right),
\end{equation}
where $\MG_a$ and $\vb_a$ are learned weights applied to the tokens $\MX$, and $\sigma$ is a nonlinear activation. 
However, we did not see significant performance gains from this input-dependent gating mechanism (see Appendix~\ref{sec:ablation-early-1}).

\myParagraph{Additional Implementation Details}
For the Early Fusion and Late Fusion approaches, we trained the corresponding models with batch size 12 and learning rate $1\e^{-4}$, using Adam optimizer.
When finetuning the LLM in Spatial Forcing, we reduce the batch size to 6 to avoid memory issues. 
For finetuning, we train the networks for 100 epochs.
For mid-training (Section~\ref{sec:exp-midtraining}), we train the corresponding networks for 10 epochs, followed by 50 epochs of finetuning on the specific task.
We train all models using an NVIDIA A100 GPU cluster with 320 GB GPU Memory. Finetuning takes 2-3 days (depending on the model), while mid-training (Section~\ref{sec:exp-midtraining}) takes around 1 week.
When finetuning \groot, we follow the standard protocol and only train the action expert initializing with pre-trained weights. 
For the Early Fusion and the Late Fusion models, we train the fusion module and the action expert and keep everything else frozen. 
For Spatial Forcing, we apply the alignment loss to layer 9 (out of 13) of the \groot LLM and only finetune the linear projection between the vision encoder and the LLM. 
Layer 9 is chosen by analogy with the OpenVLA and $\pi_0$ implementations, where the alignment loss is applied at around $70\%$ of the depth of the LLM backbone.
We also attempted to finetune the LLM in Spatial Forcing, but obtained worse results (see Appendix~\ref{sec:app-SF-finetune-LLM}).
For evaluation purposes, 
we fix the random seeds to 42 for repeatability, without any attempt at tuning the seed for performance. 
This ensures that the benchmarking assets are randomized but the testing setup is identical across VLAs.
We also fix the noise generation in the diffusion transformer of \groot to be deterministic, and use the same generation method across all models.




\setcounter{figure}{0} 
\renewcommand{\thefigure}{A.\arabic{figure}} 

\section{Appendix: Ablations} 
\label{sec:ablations}

\begin{figure*}[t!]
    \centering
    \includegraphics[width=\textwidth, trim=0mm 0mm 0mm 29mm, clip]{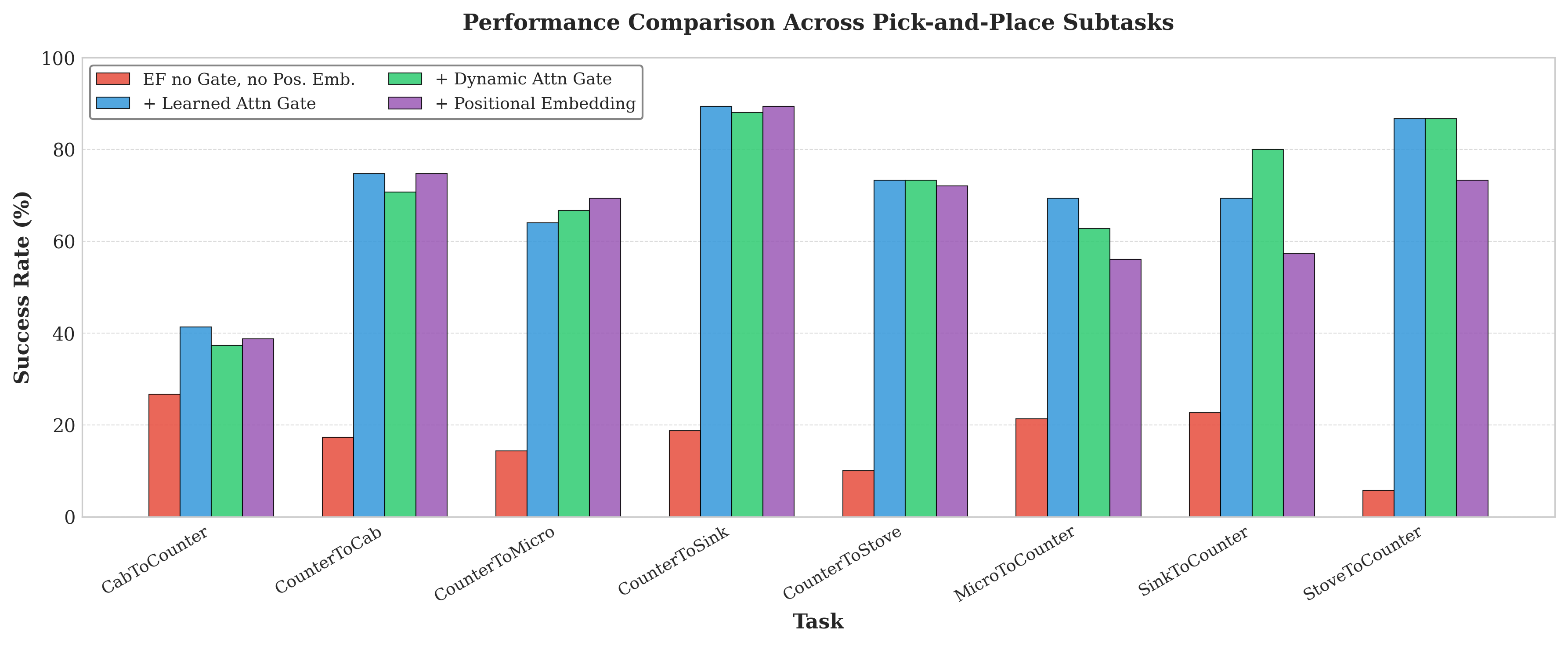}
    \caption{Ablation of fusion module within the Early Fusion (EF) strategy.}
    \label{fig:robocasa_comparison}
\end{figure*}

\subsection{Ablation of Early Fusion Architecture}
\label{sec:ablation-early-1}

This appendix provides an ablation of the main choices we made in the design of the fusion module of Section~\ref{sec:crossattn}. 
In particular, we focus on the Early Fusion model, and discuss the impact of the attention gate and positional embeddings. 
Fig.~\ref{fig:robocasa_comparison} shows that a variation of our Early Fusion model without attention gating and positional embeddings has consistently poor performance across all subtasks (5-27\% success rate). Adding a learned attention gate dramatically improves performance, with most tasks jumping to 64-89\% success rate. 
Making the attention gate input-dependent (``+Dynamic Attn Gate'' in the figure), does not lead to consistent improvements, as already mentioned in Appendix~\ref{sec:architecture_details}.
Surprisingly, even the use of positional embeddings shows mixed performance, and leads to better performance in certain tasks (\eg CounterToCab, CounterToMicrowave) but worse in others (\eg SinkToCounter, StoveToCounter). 

\subsection{Best Checkpoint Analysis for Early Fusion Approach}
\label{sec:app-best-checkpoint}

\begin{figure*}[h!]
    \centering
    \includegraphics[width=\textwidth, trim=0mm 0mm 0mm 7mm, clip]{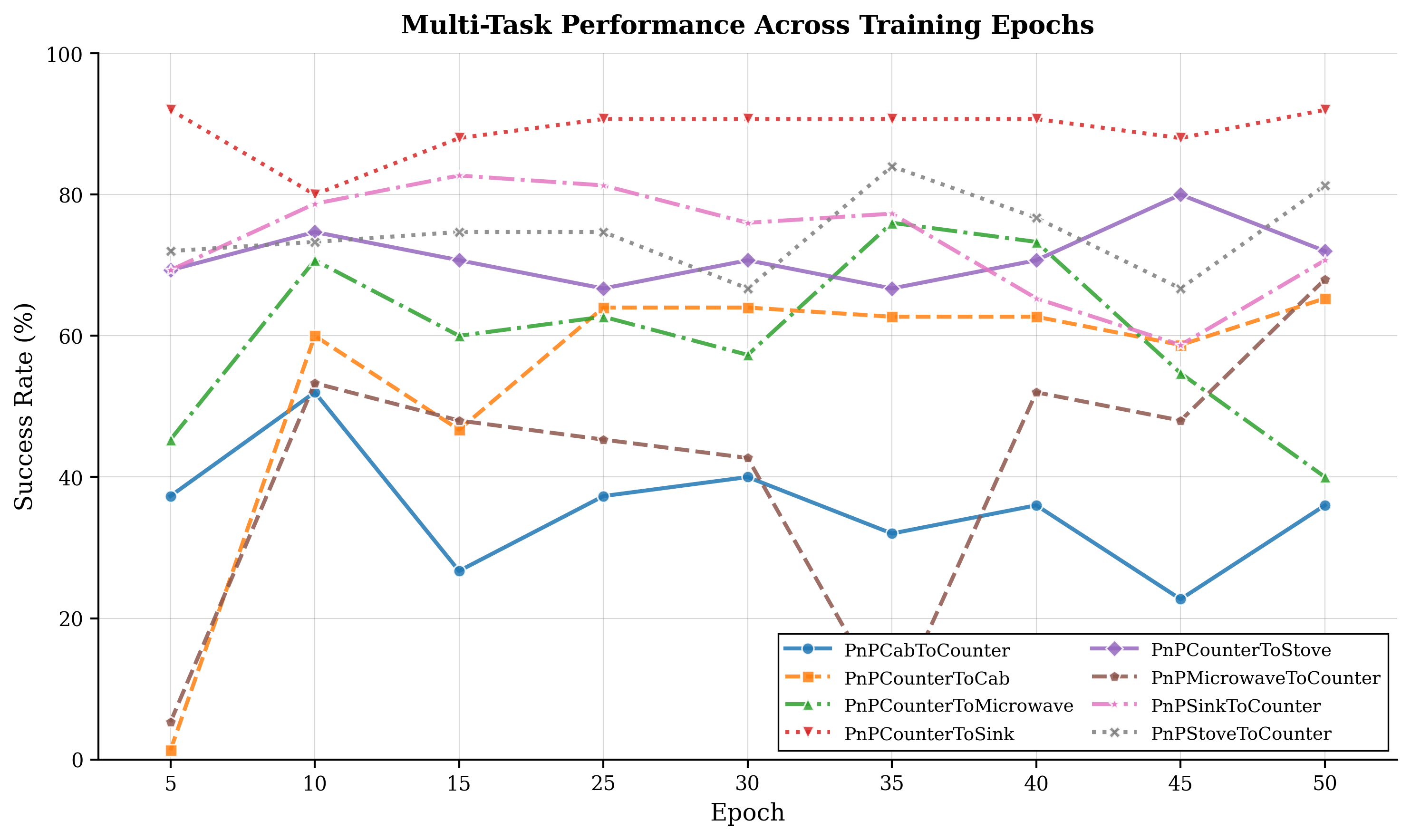}
    \caption{Early Fusion: success rate for different model checkpoints/epochs on RoboCasa. Training dynamics are task-dependent and often exhibit large fluctuations. }
    \label{fig:training_dyns}
\end{figure*}

Fig.~\ref{fig:training_dyns} shows the performance of the Early Fusion model at different stages of the training process. We observe task-specific training dynamics indicating distinct convergence patterns: for instance, PnPCounterToSink achieves 92\% success by epoch 5, while PnPCounterToCab requires 50 epochs to reach 65.3\% peak performance. We observe evidence of multi-task interference in PnPCounterToMicrowave, which peaks at 76\% (epoch 35) but degrades to 40\% by epoch 50, indicating potential catastrophic forgetting. Cabinet-related tasks consistently underperform (52-65.3\%), identifying cabinet manipulation as a key bottleneck. These results highlight the challenges of multi-task visuomotor learning and suggest that task-specific training schedules or adaptive learning rates may be necessary to optimize performance across heterogeneous manipulation tasks.

\subsection{Spatial Forcing: Finetuning LLM}
\label{sec:app-SF-finetune-LLM}


For completeness, we report results obtained with two different settings when implementing the Spatial Forcing strategy; these results guided our design for the evaluations in the main paper.
In the first setting, we apply the Spatial Forcing alignment loss to Layer 9 of the LLM and only finetune the projection module that connects the vision encoder with the LLM in \groot-N1.5. These are the results reported in the main paper.
In the second setting, we apply the alignment loss to the same LLM layer, but we also finetune the LLM in addition to the projection module. 
Table~\ref{tab:robocasaSFwithAndWithoutLLM} reports the results for the two settings across all the RoboCasa tasks.
We notice that the version that does not finetune the LLM is superior across the board (in a statistically significant way).
Intuitively, the version that finetunes the LLM has many more free parameters, and is more prone to overfitting to the training data, hindering performance.
This motivates 
our choice to focus on the first setting in the main paper. 

\begin{table*}[h!]
\centering
\scriptsize
\resizebox{\textwidth}{!}{%
\begin{tabular}{l|llllllll|l}
\hline
\textbf{Method} 
& \textbf{CabToCtr}
& \textbf{CtrToCab}
& \textbf{CtrToMicrowave}
& \textbf{CtrToSink}
& \textbf{CtrToStove}
& \textbf{MicrowaveToCtr}
& \textbf{SinkToCtr}
& \textbf{StoveToCtr} 
& \textbf{Average} \\
\hline

Spatial Forcing (no LLM ft.)
& \cellcolor{best}\textbf{29.3}
& \cellcolor{best}\textbf{68.0}
& \cellcolor{best}\textbf{66.7}
& \cellcolor{best}\textbf{76.0}
& \cellcolor{best}\textbf{72.0}
& \cellcolor{best}\textbf{60.0}
& \cellcolor{best}\textbf{84.0}
& \cellcolor{best}\textbf{90.7}
& \cellcolor{best}\textbf{68.3} \\

Spatial Forcing (LLM ft.)
& 16.0\,{\scriptsize (p=0.123)}
& 24.0\,{\scriptsize (p<0.001)}
& 26.7\,{\scriptsize (p<0.001)}
& 13.3\,{\scriptsize (p<0.001)}
& 38.7\,{\scriptsize (p<0.001)}
& 22.7\,{\scriptsize (p<0.001)}
& 53.3\,{\scriptsize (p<0.001)}
& 54.7\,{\scriptsize (p<0.001)}
& 31.2\,{\scriptsize (p<0.001)} \\

\hline
\end{tabular}}
\caption{RoboCasa results with Spatial Forcing, with and without LLM finetuning. Green indicates best result per column. $p$ values are computed against the results without LLM finetuning.\label{tab:robocasaSFwithAndWithoutLLM}}
\end{table*} 

\section{Appendix: McNemar's Statistical Test} 
\label{sec:app_mcNemar}

To assess whether the performance difference (\ie difference in the success rate) between two VLAs is statistically 
significant, we employ the two-sided McNemar's test~\cite{mcnemar1947note}, a 
statistical test for paired binary data. 
The test is \emph{two-sided} in that it evaluates whether the two methods
produce statistically different outcomes, without assuming a priori
which method performs better.
This test is appropriate because our models have binary outcomes on each trial ({\small SUCCESS/FAIL}), hence they involve \emph{binary data}; moreover,  
we evaluate both models on the same set of experiments (\ie identical testing conditions), hence we have \emph{paired observations}. 
Next, we provide a standard introduction to the test.

\myParagraph{Contingency Table Construction} 
Consider the case where we compare two VLAs, namely VLA1 and VLA2.
Let $\vxx^{(1)}$ and $\vxx^{(2)}$ be the binary vectors containing the outcomes of the tests 
for VLA1 and VLA2, respectively. For instance, $\vxx^{(1)}_i$ contains the outcome ({\small SUCCESS/FAIL}) of VLA1 on the $i$-th trial.\footnote{In RoboCasa, we have 5 episodes per task, and 15 runs per episodes, hence we have 75 trials to compute $p$ values for each task, and $600$ trials to compute $p$ values across all tasks.} 
To compute the $p$ values from the McNemar test, we start by constructing 
the $2 \times 2$ contingency table:

\begin{equation}
\begin{array}{c|cc}
 & \text{VLA2 Success} & \text{VLA2 Failure} \\
\hline
\text{VLA1 Success} & a & b \\
\text{VLA1 Failure} & c & d \\
\end{array}
\end{equation}
where $a$ is the number of trials where both VLA1 and VLA2 succeeded, 
$b$ is the number of trials where VLA1 succeeded while VLA2 failed, 
$c$ is the number of trials where VLA2 succeeded while VLA1 failed, and 
$d$ is the number of trials where they both failed.

\myParagraph{Test Statistic Computation}
McNemar's test focuses on the \emph{discordant pairs} ($b$ and $c$), which represent 
episodes where the two models disagree. Under the null hypothesis $H_0$ that both 
models have equal probability of success, we expect $b = c$.
Following standard practice, for small sample sizes ($b + c < 25$), we use the \emph{exact binomial test}:
\begin{equation}
p = 2 \sum_{i=0}^{\min(b,c)} \binom{b+c}{i} \cdot 0.5^{b+c}
\end{equation}

For larger sample sizes ($b + c \geq 25$), we use the \emph{chi-squared approximation 
with Yates' continuity correction}:
\begin{equation}
\chi^2 = \frac{(|b - c| - 1)^2}{b + c}
\end{equation}

The $p$-value is then computed as: 
\begin{equation}
p = 1 - F_{\chi^2_1}(\chi^2)
\end{equation}
where $F_{\chi^2_1}$ is the cumulative distribution function of the chi-squared 
distribution with 1 degree of freedom. A small $p$ value indicates that the two methods differ significantly in their success rates on the paired trials.
 We return $p=1.0$ when $b+c=0$ (no discordant pairs, \ie identical performance).

\myParagraph{Aggregated Analysis}
For overall model comparison across multiple tasks, we aggregate all task-level 
success/failure outcomes and apply McNemar's test to the combined contingency table, 
providing a single $p$ value for the overall performance difference.


\section{Appendix: Taming the Randomness} 
\label{sec:app_randomness}

New papers often claim state-of-the-art performance from small increments in the success rate.
 In this appendix, we show that the average success rate alone might be a poor indicator of improved performance, since small increments
 are sometimes hard to distinguish from stochasticity. Our results reinforce recent efforts in the community (\eg~\cite{lbmtri2025,Snyder25arxiv-isYourPolicyBetter}) to push for higher standards in VLA 
 experimentation, based on more rigorous statistical analysis.

More in detail, this appendix shows that even when (i) removing all sources of randomness in the benchmarking infrastructure (\ie fixing randomization seeds such that the simulation setup is identical across different VLAs), and (ii) increasing the number of trials beyond what is commonly done in the literature, the mean success rates have non-negligible fluctuations 
when repeating the same experiment multiple times. Intuitively, even when the testing setup is  fixed,  the randomization in the diffusion transformer in the VLA's action expert causes relatively large changes in the mean success rate.
This observation has two implications. 
First, one can get widely different results even in identical testing conditions and ---even when fixing the randomization seed for the action expert--- the choice of the seed is consequential for the evaluation.
Second, one should be careful in drawing conclusions just by looking at mean success rate (as done by most VLA papers in the literature), since the success rate largely varies even when repeating the same experiment.

\begin{table}[h!]
\centering
\setlength{\tabcolsep}{3.6pt}
\scriptsize
\begin{tabular}{l|cccccc|cccccc}
\hline
 & \multicolumn{6}{c|}{\textbf{Epoch 20}} & \multicolumn{6}{c}{\textbf{Epoch 80}} \\
\textbf{Trials per episode}
 & 10 & 20 & 30 & 40 & 50 & 100
 & 10 & 20 & 30 & 40 & 50 & 100 \\
\hline

\textbf{Mean}
& 38.2 & 38.5 & 39.1 & 38.6 & 39.0 & 38.4
& 20.6 & 20.8 & 21.7 & 19.9 & 20.0 & 21.2 \\
\hline

\textbf{Std. Dev.}
& 2.2 & 1.2 & 1.0 & 1.1 & 0.8 & 0.8
& 4.8 & 2.6 & 2.1 & 1.9 & 1.7 & 1.0 \\

\textbf{Min}
& 34 & 36 & 37.3 & 36.5 & 37.6 & 37
& 12 & 15 & 19.3 & 16 & 17.2 & 19.6 \\

\textbf{Max}
& 42 & 40 & 40 & 40 & 40 & 39.6
& 28 & 25 & 25.3 & 22 & 23.2 & 22.6 \\

\hline
\end{tabular}
\caption{Success rate statistics across 10 repetitions of the same experiment, for different numbers of trials for \groot-N1.5 at Epoch 20 and Epoch 80.}
\label{tab:trial_stats_transposed}
\end{table}

To investigate the impact of the action expert randomness on the average success rate, for each RoboCasa episode, we use the same scene across trials and we do not fix the noise in the diffusion process of the action expert.  
We focus on the PnPCabToCounter RoboCasa task and we evaluate two checkpoints of \groot-N1.5 (epoch 20 and epoch 80) to understand if the randomness in the action expert behavior decreases with more training iterations. For each checkpoint, we repeat 10, 20, 30, 40, 50, and 100 trials for each episode and use the trials to compute the overall success rate.
Then, we repeat the same experiment 10 times to assess variations in the success rate.
Table~\ref{tab:trial_stats_transposed} reports, for each epoch and for each number of trials, the mean success rate, its standard deviation, as well as 
the minimum and maximum success rate. 

First of all, we note that for the number of trials typically done in related work (10-20, \cf~\cite{bjorck2025gr00t} in the main paper), there is a very large fluctuation in the 
experimental results, with a standard deviation of 4.8 at Epoch 80; this translates into fluctuations of 8-10\% in mean success rate across identical experiments, while often
related work claims performance advantages from more modest increases in success rates. 
Second, while the success rate variability decreases with more runs, the action expert randomness is still responsible for fluctuations in the order of 2-3\%. 
Note that performing 100 trials per episode, leads to 4000 overall experiments across all RoboCasa tasks, which requires substantial computational resources.
Third, the checkpoint at Epoch 20 consistently outperformed checkpoint Epoch 80 (37-39.6\% vs. 19.6-22.6\% success rate); while this is not surprising 
from the analysis we provided in Appendix~\ref{sec:app-best-checkpoint}, it is interesting to notice that weaker models (Epoch 80) show higher success rate 
variations overall.

In summary, the randomness in the success rate of VLAs  based on diffusion policies cannot be fully tamed by just increasing the number of trials, even assuming one has access
to enough computational resources. Therefore, the mean success rate cannot be taken at face value as the only indicator when comparing two VLAs. 
The path we took in this paper to address this concern is twofold: (i) we fixed randomization seeds and used deterministic noise in the action expert to ensure repeatability
(and without any effort to tune them for performance),  and (ii) we computed $p$ values using McNemar's test (Appendix~\ref{sec:app_mcNemar}) to assess significance levels of the test results.

\section{Appendix: Additional Experiments} 
\label{sec:extra_experiments}

\subsection{LIBERO Results}

In this section, we complement Section~\ref{sec:exp-fusion} with additional experiments on the {\small LIBERO} benchmark.
Table~\ref{tab:liberoResults} summarizes the  results of our evaluation of the \groot baseline and the implemented geometric VLAs. We do not test on {\small LIBERO-GOAL} due to a known issue with \groot on this benchmark\footnote{See https://github.com/huggingface/lerobot/issues/2457}.
The average success rates in Table~\ref{tab:liberoResults} seem to suggest that Late Fusion and Spatial Forcing outperform the baseline, while Early Fusion lags slightly behind.
However, from a closer inspection of the $p$ values ---and consistently with Section~\ref{sec:exp-fusion}--- the differences between the baseline and the geometric VLAs are not statistically significant and are likely the result of random fluctuations rather than actual performance gain or loss.

\begin{table}[h!]
\centering
\scriptsize
\resizebox{\textwidth}{!}{%
\begin{tabular}{l|llll|l}
\toprule
& \textbf{Spatial}
& \textbf{Object}
& \textbf{LIBERO-10}
& \textbf{LIBERO-90}
& \textbf{Average} \\
\midrule

\groot-N1.5
& \cellcolor{best}96.7
& 95.3
& 78.0
& 81.8
& 87.9 \\

Early Fusion
& 94.0\,{\scriptsize (p=0.289)}
& 94.0\,{\scriptsize (p=0.804)}
& 76.0\,{\scriptsize (p=0.755)}
& 82.2\,{\scriptsize (p=0.307)}
& 86.6\,{\scriptsize (p=0.138)} \\

Late Fusion
& 93.3\,{\scriptsize (p=0.227)}
& \cellcolor{best}96.0\,{\scriptsize (p=1.000)}
& 83.3\,{\scriptsize (p=0.230)}
& \cellcolor{best}91.1\,{\scriptsize (p=0.804)}
& 90.9\,{\scriptsize (p=0.561)} \\

Spatial Forcing
& 95.3\,{\scriptsize (p=0.727)}
& \cellcolor{best}96.0\,{\scriptsize (p=1.000)}
& \cellcolor{best}84.7\,{\scriptsize (p=0.144)}
& 90.0\,{\scriptsize (p=1.000)}
& \cellcolor{best}91.5\,{\scriptsize (p=0.295)} \\

\bottomrule
\end{tabular}
}
\caption{LIBERO benchmark results. Green indicates best result per column; yellow indicates second best. $p$ values are computed against \groot-N1.5.}
\label{tab:liberoResults}
\end{table}

\subsection{Impact of Training Data Scaling: Mid-training \groot-N1.5}

In this section, we present a variation of the experiments from Section~\ref{sec:exp-midtraining}, where we compare the mid-trained geometric VLAs against a mid-trained version of the \groot baseline.
This is important to confirm that the advantage observed in the geometric VLAs is actually due to the use of VGGT rather than information leakage results from the mid-training.

\myParagraph{RoboCasa}
Table~\ref{tab:midtrained_transposed_with_depth} reports the results on the RoboCasa results.
The results confirm that (i) the performance of the Early Fusion approach significantly improves thanks to mid-training even with respect to a mid-trained version of the baseline, and that (ii) the advantage becomes more statistically significant even when compared to the mid-trained \groot baseline.
In the table, we also report, for each task, average depth metrics, including RMSE, $\delta_1$, and Mean Absolute Error (MAE). However, due to task differences, it is not immediate to correlate VGGT depth prediction quality with success rate in this case (\cf the more insightful visualization in Fig.~\ref{fig:seven_panel_layout}).

\begin{table*}[h!]
\centering
\scriptsize
\resizebox{\textwidth}{!}{%
\begin{tabular}{l|llllllll|l}
\hline
\textbf{Method} 
& \textbf{CabToCtr}
& \textbf{CtrToCab}
& \textbf{CtrToMicrowave}
& \textbf{CtrToSink}
& \textbf{CtrToStove}
& \textbf{MicrowaveToCtr}
& \textbf{SinkToCtr}
& \textbf{StoveToCtr} 
& \textbf{Average} \\
\hline



GR00T (mid-trained)
& 50.7
& \cellcolor{best}72.0
& \cellcolor{best}80.0
& \cellcolor{best}100.0
& \cellcolor{best}81.3
& 61.3
& 66.7
& 65.3
& 72.2 \\

Early Fusion (mid-trained)
& \cellcolor{best}52.0\,{\scriptsize (p=1.000)}
& \cellcolor{best}72.0\,{\scriptsize (p=1.000)}
& 69.3\,{\scriptsize (p=0.134)}
& 94.7\,{\scriptsize (p=0.125)}
& 80.0\,{\scriptsize (p=1.000)}
& \cellcolor{best}68.0\,{\scriptsize (p=0.424)}
& \cellcolor{best}81.3\,{\scriptsize (p=0.063)}
& \cellcolor{best}84.0\,{\scriptsize (p=0.004)}
& \cellcolor{best}75.2\,{\scriptsize (p=0.168)} \\

\hline

\textbf{Depth RMSE}
& 0.114±0.035
& 0.094±0.025
& 0.114±0.041
& 0.064±0.026
& 0.085±0.038
& 0.082±0.011
& 0.069±0.024
& 0.084±0.035
& 0.088±0.035 \\

\textbf{Depth $\delta_1$}
& 0.830±0.062
& 0.869±0.041
& 0.829±0.060
& 0.927±0.021
& 0.894±0.061
& 0.869±0.034
& 0.925±0.033
& 0.873±0.062
& 0.877±0.060 \\

\textbf{Depth MAE}
& 0.068±0.027
& 0.056±0.021
& 0.071±0.029
& 0.037±0.017
& 0.057±0.035
& 0.048±0.009
& 0.038±0.018
& 0.056±0.035
& 0.054±0.028 \\

\hline
\end{tabular}}
\caption{Performance of mid-trained VLAs on RoboCasa. Green indicates best result per column; yellow indicates second best. For each task, we also report average depth metrics, including  RMSE, $\delta_1$ score, and the Mean Absolute Error (MAE).\label{tab:midtrained_transposed_with_depth}}
\end{table*}



\myParagraph{{\small LIBERO}} 
Table~\ref{tab:liberoResults_midtrained} shows mid-training results for {\small LIBERO}, complementing the results in Section~\ref{sec:exp-midtraining}. Consistently with RoboCasa, we observe that after mid-training, the success rate of the Early Fusion approach increases.
At the same time, the $p$ value for the overall experiment remains large, hence we cannot draw a very strong conclusion in this case. 

\begin{table}[h!]
\centering
\scriptsize
\resizebox{\textwidth}{!}{%
\begin{tabular}{l|llll|l}
\toprule
& \textbf{Spatial}
& \textbf{Object}
& \textbf{LIBERO-10}
& \textbf{LIBERO-90}
& \textbf{Average} \\
\midrule



\groot-N1.5 
& 96.7
& \cellcolor{best}95.3
& \cellcolor{best}78.0
& 81.8
& \cellcolor{best}87.9 \\

Early Fusion (mid-trained)
& \cellcolor{best}98.0\,{\scriptsize (p=0.727)}
& 94.7\,{\scriptsize (p=1.000)}
& 76.0\,{\scriptsize (p=0.735)}
& \cellcolor{best}82.2\,{\scriptsize (p=0.327)}
& 87.7\,{\scriptsize (p=0.440)} \\

\bottomrule
\end{tabular}
}
\caption{Performance of mid-trained VLAs on {\small LIBERO}. Green indicates best result.}
\label{tab:liberoResults_midtrained}
\end{table}

%






\subsection{Inspecting the Attention Masks in Early Fusion}
\begin{figure*}[h!]
    \centering
    \includegraphics[width=10cm]{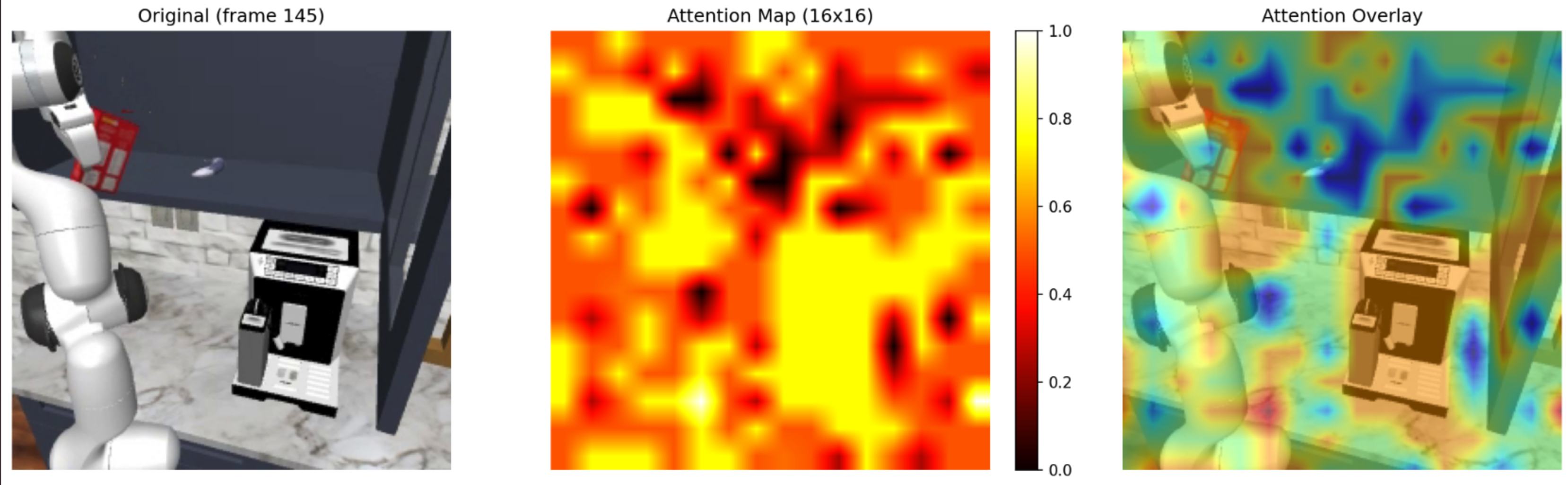}
    \caption{Early Fusion: RGB image, attention mask, and attention mask overlaid on original image for a sample input of the RoboCasa dataset.}
    \label{fig:attention_mask}
\end{figure*}

Fig.~\ref{fig:attention_mask} shows the original RGB observations and the corresponding attention masks produced by the Early Fusion model on a sample image from the RoboCasa benchmark. 
While the mask is fairly noisy, higher attention values tend to concentrate around the robot arm and end-effector regions. This hints to the fact that the model learns spatially-aware representations and captures task-relevant geometric structure for manipulation. 

\subsection{Linear Probing for Normals Estimation}
\label{app:normalProbing}

This appendix provides extra results to further support our findings in Section~\ref{sec:probing}. In that section, we argue that Early and Late Fusion allow injecting geometric information into a VLA, by showing that the resulting architectures are able to predict monocular depth via linear probes, while the \groot-N1.5 baseline falls short in predicting accurate depth.
However, one might argue that predicting depth is only one aspect of geometric understanding, and other aspects are also important for manipulation. This section shows that our findings hold even when predicting other types of geometric information, namely \emph{surface normals}.

\begin{table}[t]
\centering
\begin{tabular}{lccc}
\hline
 & Mean Error ($\downarrow$) & Median Error ($\downarrow$) & Within $30^\circ$ ($\uparrow$) \\
\hline
GR00T - vision encoder probe & 44.97 & 45.26 & 0.33 \\
GR00T - VLM probe            & 44.43 & 44.15 & 0.34 \\
VGGT probe                   & {\bf 39.62} & {\bf 33.86} & {\bf 0.45} \\
\hline
Early Fusion probe                & 41.29 & 34.40 & 0.44 \\
Late Fusion probe                & 42.34 & 38.96 & 0.39 \\
\hline
\end{tabular}
\caption{Surface normal probing. Mean and median angular errors between predicted and ground truth normals and fraction of normals with angular error $\leq 30^\circ$.
 \label{tab:normal_estimation_results}}
\end{table}

Surface normal estimation is a fundamental capability for robotic manipulation because manipulation depends on making physically meaningful contact with object surfaces. While object pose and depth indicate where objects are located, surface normals describe how a robot can interact with them by encoding local contact geometry and feasible force directions. Normals are essential for grasp synthesis, since stable grasps often require alignment with opposing or complementary surface orientations. 
In this sense, surface normal estimation forms a key bridge between visual perception and physical interaction. 

\myParagraph{Probing Setup} We follow the same probing protocol as Section~\ref{sec:probing}, but now we train the linear probe to predict a 3D vector (surface normal) for each pixel in the input image instead of a scalar depth value. 
We train the linear probe for 10 epochs on the NYU Depth V2 dataset. We use a cosine similarity loss; more precisely, we minimize $1 - \vn\tran_{\text{pred}} \vn_{\text{gt}}$, where $\vn_{\text{pred}}$ is the predicted normal at a pixel and $\vn_{\text{gt}}$ is the ground truth.  To evaluate probing performance, we compute ---for each pixel--- the angle (in degrees) between the estimated and ground truth normal, and then report the mean and median error ($\downarrow$),  as well as fraction of pixels with angular error less than $30^\circ$ ($\uparrow$), across the validation set. 

\myParagraph{Probing Results} The results in Table~\ref{tab:normal_estimation_results} confirm our findings in Section~\ref{sec:probing}. VGGT achieves top performance across the spectrum: as expected, the \GFM has a good geometric understanding of the image and can perform surface normal prediction, even when using a simple linear probe. On the other hand, the original \groot-N1.5 VLA achieves much higher errors, both when probed after the vision encoder, as well as when probed after the VLM, consistently with the results in Section~\ref{sec:probing}. Finally, the Early and Late Fusion architectures allow to inject geometric information into the VLA, reducing the normal estimation errors. In particular, the Early Fusion probe achieves results close to VGGT, leading to a similar conclusion as the one we provided in Section~\ref{sec:probing}.

\subsection{Invariance to Object Appearance}

\begin{figure*}[h!]
    \centering
    \includegraphics[width=\textwidth, trim=0mm 95mm 110mm 0mm, clip]{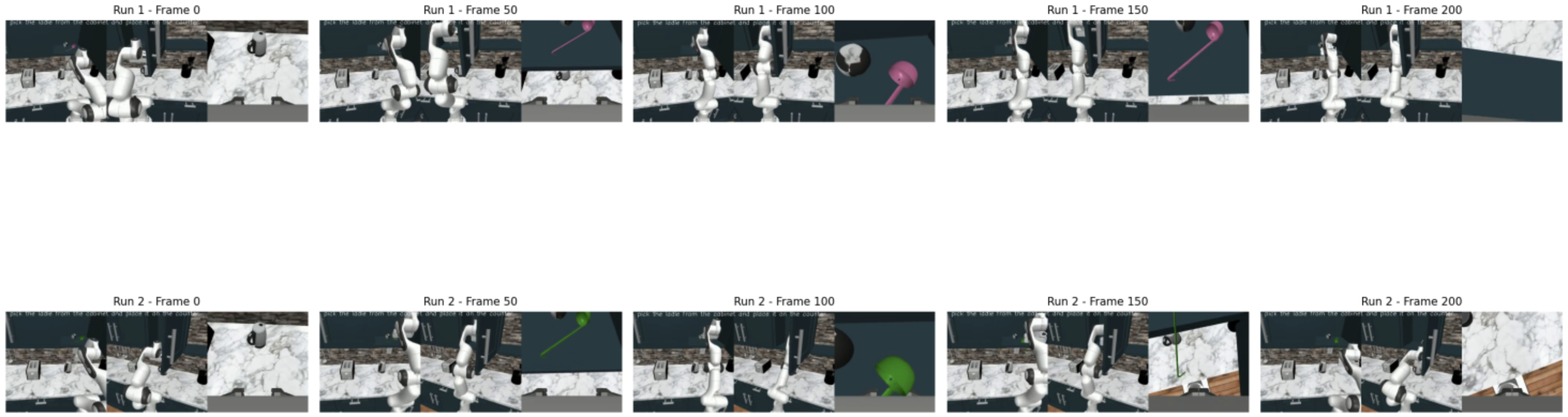}
    \includegraphics[width=\textwidth, trim=0mm 0mm 110mm 95mm, clip]{figures/rollouts.png}
    \caption{RoboCasa: object appearance variations to test VLA invariance to object color. The figure shows the images from the 3 cameras (left, right, wrist) at different time steps (0, 50, 100, 150) for two trials with randomized object appearance.  Top (Run 1): Ladle rendered in magenta. Bottom (Run 2): Ladle rendered in green.}
    \label{fig:robocasa_full}
\end{figure*}

In this appendix, we report additional results that were not discussed in the main paper due to space constraints but that shed light on the models' sensitivity to object appearance.
In particular, we test the invariance of the Early Fusion approach, as well as the \groot-N1.5 baseline, to object appearance. 
Intuitively, we want to test if the use of geometry in Early Fusion gives the approach extra invariance to the object appearance, which is typically a nuisance and irrelevant 
for the task.\footnote{This is not true in general, since the task might involve appearance (\eg ``pick-up the \emph{red} mug''), but it is true on the RoboCasa benchmark, where 
objects to manipulate are specified by semantics, rather than appearance.}
To evaluate this hypothesis, we randomize object appearances in the RoboCasa benchmark while keeping the underlying geometry fixed. More precisely, in the main paper, we used the standard RoboCasa setup, where, for each episode, we repeat $N$ trials ($N=15$ in our case) where the appearance of all assets is randomized. 
However, in this section, we apply a slight modification of the RoboCasa pipeline and we only randomize the appearance of the target object across the $N$ trials, while keeping all the remaining assets fixed. 
Sample images showing the object appearance variations are given in Fig.~\ref{fig:robocasa_full}.  

Using this setup, Table~\ref{tab:robocasaRandomizedObjects} reports results for both the Early Fusion approach and the \groot baseline across the RoboCasa tasks.

\begin{table*}[h!]
\centering
\scriptsize
\resizebox{\textwidth}{!}{%
\begin{tabular}{l|llllllll|l}
\hline
\textbf{Method} 
& \textbf{CabToCtr}
& \textbf{CtrToCab}
& \textbf{CtrToMicrowave}
& \textbf{CtrToSink}
& \textbf{CtrToStove}
& \textbf{MicrowaveToCtr}
& \textbf{SinkToCtr}
& \textbf{StoveToCtr} 
& \textbf{Average} \\
\hline

\groot-N1.5 (mid-trained)
& 29.3
& \cellcolor{best}\textbf{78.7}
& \cellcolor{best}\textbf{89.3}
& 97.3
& 82.7
& \cellcolor{best}\textbf{65.3}
& 60.0
& \cellcolor{best}\textbf{92.0}
& 74.3 \\

Early Fusion (mid-trained)
& \cellcolor{best}\textbf{48.0}\,{\scriptsize (p=0.026)}
& 76.0\,{\scriptsize (p=0.625)}
& 88.0\,{\scriptsize (p=1.000)}
& \cellcolor{best}\textbf{98.7}\,{\scriptsize (p=1.000)}
& \cellcolor{best}\textbf{89.3}\,{\scriptsize (p=0.302)}
& 56.0\,{\scriptsize (p=0.210)}
& \cellcolor{best}\textbf{84.0}\,{\scriptsize (p=0.004)}
& 73.3\,{\scriptsize (p<0.001)}
& \cellcolor{best}\textbf{76.7}\,{\scriptsize (p=0.279)} \\
\hline
\end{tabular}}
\caption{RoboCasa with randomized object appearance. Green indicates best result.}
\label{tab:robocasaRandomizedObjects}
\end{table*} 


The quantitative results in Table~\ref{tab:robocasaRandomizedObjects} support empirically that (i) the advantage of the Early Fusion approach is retained also in this case, but (ii) the relative advantage between Early Fusion and baseline remains similar to the original testing setup (\cf Table~\ref{tab:midtrained_transposed_with_depth}). Hence the object appearance variations do not pose a particular challenge for neither the baseline nor the Early Fusion approach.  
This was somewhat expected since the vision encoder in both approaches abstracts low-level pixel appearance into a suitable feature space, which is less sensitive to appearance
variations by design.

\end{document}